\documentclass[journal]{IEEEtran}

\ifCLASSINFOpdf
\else
\fi
\usepackage{cite}
\usepackage[colorlinks, citecolor=black, linkcolor=black]{hyperref}
\usepackage[capitalize]{cleveref}
\usepackage{times}
\usepackage{epsfig}
\usepackage{graphicx}
\usepackage{amsmath}
\usepackage{amssymb}
\usepackage{color}
\usepackage{multirow}
\usepackage{array}
\usepackage{epstopdf}
\usepackage{subfig}
\usepackage{caption}
\usepackage[font=small]{caption}
\usepackage{booktabs}
\usepackage{enumerate}
\usepackage{float}

\newcommand{\etal}{\textit{et al}. }

\newcommand{\eg}{\textit{e.g}.}
\newcommand{\ie}{\textit{i.e}.}
\newcommand{\etc}{\textit{etc}}

\begin{document}

\title{Interpretable Detail-Fidelity Attention Network for Single Image Super-Resolution}

\author{Yuanfei~Huang,
	Jie~Li,
	Xinbo~Gao,~\IEEEmembership{Senior Member,~IEEE,}
	Yanting~Hu,
	and~Wen~Lu
\thanks{
Yuanfei Huang, Jie Li and Wen Lu are with the Video and Image Processing System Laboratory, School of Electronic Engineering, Xidian University, Xi'an 710071, China (e-mail: \href{mailto:yf_huang@stu.xidian.edu.cn}{{\color{black}yf\_huang@stu.xidian.edu.cn}}; leejie@mail.xidian.edu.cn; luwen@mail.xidian.edu.cn).

Yanting Hu is with the School of Medical Engineering and Technology, Xinjiang Medical University, Urumqi 830011, Xinjiang, P. R. China (e-mail: yantinghu2012@gmail.com)

Xinbo Gao is with the School of Electronic Engineering, Xidian University, Xi¡¯an 710071, China (e-mail: xbgao@mail.xidian.edu.cn) and with the Chongqing Key Laboratory of Image Cognition, Chongqing University of Posts and Telecommunications, Chongqing 400065, China (e-mail: gaoxb@cqupt.edu.cn).

\IEEEcompsocthanksitem (Corresponding author: Xinbo Gao.)}}

%



\maketitle

\begin{abstract}
Benefiting from the strong capabilities of deep CNNs for feature representation and nonlinear mapping, deep-learning-based methods have achieved excellent performance in single image super-resolution. However, most existing SR methods depend on the high capacity of networks which is initially designed for visual recognition, and rarely consider the initial intention of super-resolution for detail fidelity. Aiming at pursuing this intention, there are two challenging issues to be solved: (1) learning appropriate operators which is adaptive to the diverse characteristics of smoothes and details; (2) improving the ability of model to preserve the low-frequency smoothes and reconstruct the high-frequency details.
To solve them, we propose a purposeful and interpretable detail-fidelity attention network to progressively process these smoothes and details in divide-and-conquer manner, which is a novel and specific prospect of image super-resolution for the purpose on improving the detail fidelity, instead of blindly designing or employing the deep CNNs architectures for merely feature representation in local receptive fields. Particularly, we propose a Hessian filtering for interpretable feature representation which is high-profile for detail inference, a dilated encoder-decoder and a distribution alignment cell to improve the inferred Hessian features in morphological manner and statistical manner respectively.
Extensive experiments demonstrate that the proposed methods achieve superior performances over the state-of-the-art methods quantitatively and qualitatively. Code is available at \href{https://github.com/YuanfeiHuang/DeFiAN}{github.com/YuanfeiHuang/DeFiAN}.
\end{abstract}


%
\IEEEpeerreviewmaketitle

\section{Introduction}
\IEEEPARstart{S}{ingle} image super-resolution (SISR), aiming at reconstructing a high-resolution (HR) image from a single low-resolution (LR) image obtained by limited imaging devices, is a representative branch of low-level vision and widely used in computer vision applications where high-frequency details are greatly desired, \eg, medical imaging, security and surveillance.
For decades, numerous researches have been proposed to solve this ill-posed problem of SR, including interpolation-based~\cite{Bicubic1981TASSP}, reconstruction-based~\cite{MarquinaA2008JSC,DongW2011TIP} and example-learning-based~\cite{YangJ2010TIP,ZeydeR2010,HuY2016TIP,HuangY2018TIP} methods.
Recently, with the development of the high-profile deep convolutional neural network (CNN), deep-learning-based SR methods have achieved numerous attentions as its excellent performance and real-time processing.

\begin{figure}
    \vspace{-0.2cm}
	\begin{minipage}[b]{0.95\linewidth}
		\subfloat[``{\em img086}'' in Urban100]{
		\centering
	       \includegraphics[width=0.5\linewidth]{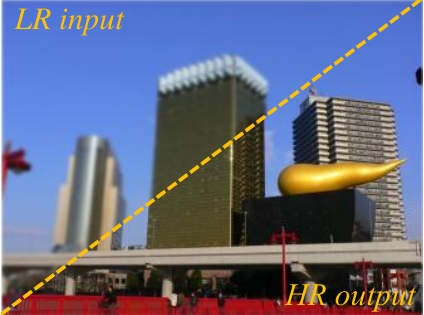}
		}
		\subfloat[scaled Hessian features]{
			\centering
			\includegraphics[width=0.5\linewidth]{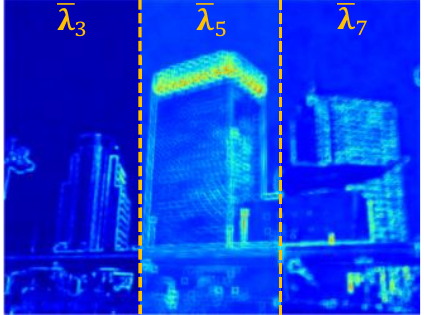}
		}\\
		\subfloat[residual image]{
			\centering
			\includegraphics[width=0.5\linewidth]{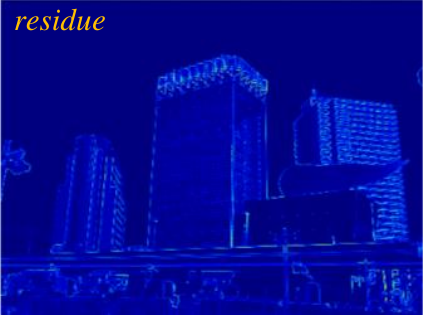}
		}
        \subfloat[detail-fidelity attention $\boldsymbol{\bar a}$]{
			\centering
			\includegraphics[width=0.5\linewidth]{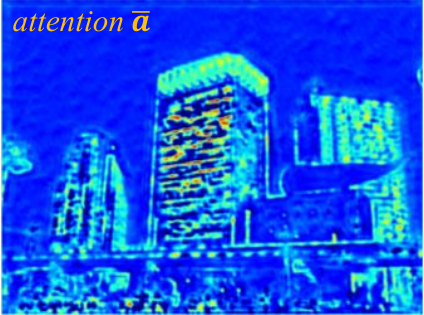}
		}
	\end{minipage}
	\caption{Visualization of attentions in 5-th {\em DeFiAM} modules (from $\textit{DeFiAN}_S$). Specifically, $\boldsymbol{\bar{\lambda}}_{3}$, $\boldsymbol{\bar{\lambda}}_{5}$ and $\boldsymbol{\bar{\lambda}}_{7}$ denote the average map of scaled Hessian features when $ker$=3, 5 and 7. The attention map $\boldsymbol{\bar{a}}$ is inferred from these Hessian features and covers the residues of image well.}
    \vspace{-0.6cm}
	\label{fig:visualization_hessian}
\end{figure}
Initially, Dong \etal\cite{DongC2016TPAMI} proposed a super-resolution convolutional neural network (SRCNN) by stacking a shallow CNN to learn the nonlinear LR-to-HR mappings, which outperforms most conventional example-learning-based SR methods.
To facilitate the training phase and eliminate the vanishing/exploding gradients problem, Kim \etal further proposed the accurate VDSR~\cite{KimJ2016CVPR_VDSR} by stacking more than 20 convolutional layers with global residual learning. However, there also exists gradient vanishing/exploding by only using the global residual learning when the depth of network increases. Inspired by ResNet~\cite{HeK2016CVPR}, the local residual learning was applied into SR task, \eg, Tai \etal proposed DRRN~\cite{TaiY2017CVPR} by utilizing the multi-path local residual learning, Ledig \etal proposed SRResNet~\cite{LedigC2017CVPR} by stacking the post-activated residual blocks to build very deep SR backbone, Lim \etal proposed EDSR~\cite{LimB2017CVPRW} to obtain better performance by removing all the BN operations in SRResNet and achieved excellent performances.
Furthermore, aiming at fully using the information flow of intermediate features in network, densely connection of DenseNet~\cite{HuangG2017CVPR} was applied into SR task as SRDenseNet~\cite{TongT2017ICCV}, MemNet~\cite{TaiY2017ICCV}, RDN~\cite{ZhangY2018CVPR} and \etc.
Besides, the feature diversity and informativeness in each layer is also an important issue for improving performance, thus, attention mechanism was highlighted and widely applied for feature enhancement, such as channel-wise attention in SENet~\cite{HuJ2018CVPR}, spatial-wise attention in CBAM~\cite{WooS2018ECCV}, non-local attention in \cite{WangX2018CVPR}, and further applied into the corresponding SR tasks: RCAN~\cite{ZhangY2018ECCV}, CSFM~\cite{HuY2019TCSVT} and RNAN~\cite{ZhangY2019ICLR}.

However, higher computational complexities and numerous parameters are introduced as the networks become deeper and wider, then several SISR methods were proposed to reduce them. For example, aiming at reducing parameters, recursive learning was exploited by stacking deep convolution layers with weight sharing, which recursively calls a single block throughout the whole network~\cite{KimJ2016CVPR_DRCN, TaiY2017CVPR}. Meanwhile, to reduce the computational complexities in model training and inference, lightweight architectures was designed using information diffluence, \eg, IDN~\cite{HuiZ2018CVPR}, MSRN~\cite{LiJ2018ECCV} and CARN~\cite{AhnN2018ECCV}. Moreover, Dong \etal\cite{DongC2016ECCV} and Shi \etal\cite{ShiW2016CVPR} respectively utilized the transposed convolution and sub-pixel convolution module to upscale the inferred features in tail and limited the input of LR size to alleviate the computational loads. For larger upscaling factor, Lai \etal\cite{LaiW2017CVPR} proposed a Laplacian pyramid network for super-resolution (LapSRN) via reconstructing the sub-band residual HR images at multiple pyramid levels.

Nevertheless, the existing SR methods almost consider the information in local receptive fields where only $3\times3$ kernels of convolution are utilized to represent the local consistent details of feature. However, the collected LR images are full of low-frequency smoothes and high-frequency details, thus, it is natural to raise two issues: (1) It is difficult to learn a perfect convolutional operator, which is adaptive to the diverse characteristics of smoothes and details; (2) How to improve the ability to preserve the low-frequency smoothes and reconstruct the high-frequency details?

(1) For the first issue, since the low-frequency smoothes and high-frequency details have different characteristics of representation, and the ill-posed problem of SR is more sensitive to the fidelity of deficient details, it is better to solve it in a divide-and-conquer manner.

(2) For the second issue, following the first issue, we should preserve the low-frequency smoothes and reconstruct the high-frequency details as better as possible, which aims at reconstructing the residues (in architectures with global residual learning) using detail-fidelity features as in Fig.~\ref{fig:visualization_hessian}.

\begin{figure*}
	\centering
	\includegraphics[width=1\linewidth]{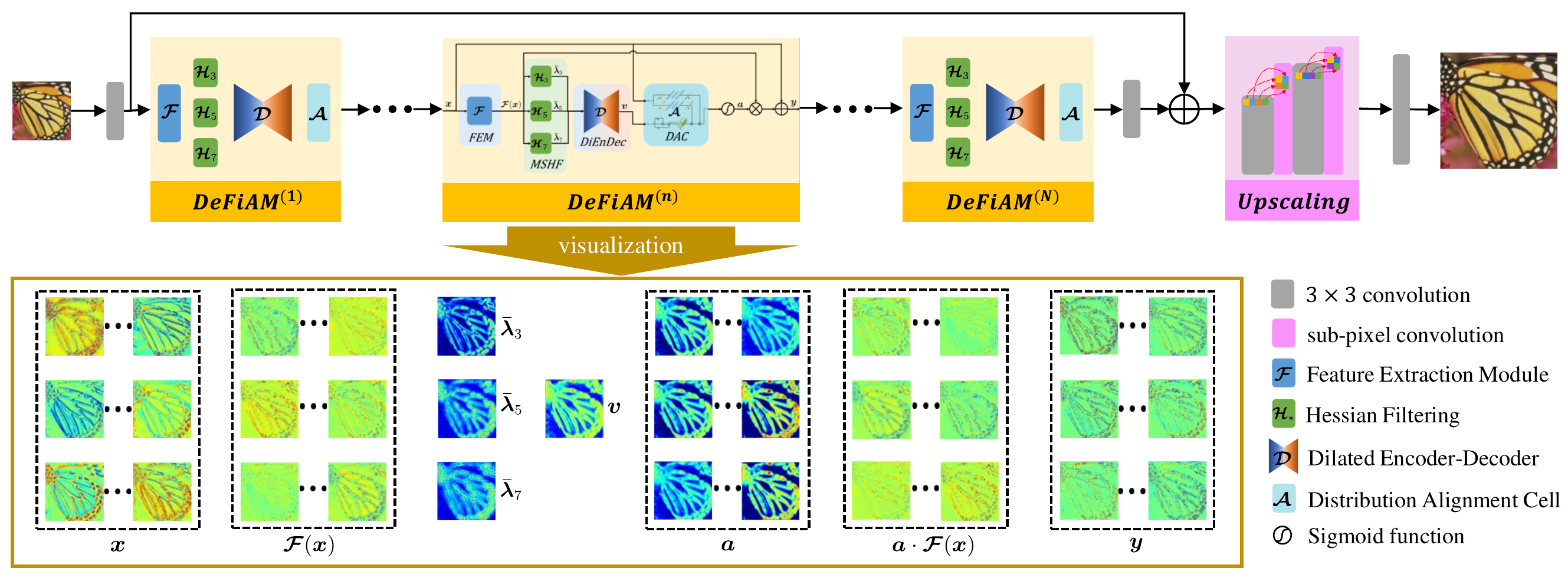}
	\caption{Architecture of the proposed method. There are $N$ stacked {\em DeFiAM} modules for progressively improving the detail fidelity in size of low-resolution, specifically, the residual branch is a chain of $M$ residual channel attention blocks for representing the local residual features as~\cite{ZhangY2018ECCV}, and an attention branch (including {\em MSHF}, {\em DiEnDec} and {\em DAC}) is designed to learn the attention on detail fidelity.}
	\label{fig:DeFiAN}	
    \vspace{-0.2cm}
\end{figure*}

For these issues and to process the low-frequency smoothes and high-frequency details in a divide-and-conquer manner, we propose a purposeful and interpretable method to improve SR performance using Detail-Fidelity Attention in very deep Networks ({\em DeFiAN}), as Fig.~\ref{fig:DeFiAN} shows. The major contributions of the proposed method are:
\begin{itemize}
\item Introducing a detail-fidelity attention mechanism in each module of networks to adaptively improve the desired high-frequency details and preserve the low-frequency smoothes in a divide-and-conquer manner, which is purposeful for SISR task.

\item Proposing a novel multi-scale Hessian filtering ({\em MSHF}) to extract the multi-scale textures and details with the maximum eigenvalue of scaled Hessian features implemented using high-profile CNNs. Unlike the conventional CNN features in most existing SR methods, the proposed {\em MSHF} is interpretable and specific to improve detail fidelity. Besides, the proposed multi-scale and generic Hessian filtering are the first attempts for interpretable detail inference in SISR task, and could be implemented using GPU-accelerate CNNs without any calculation of intricate inverse of matrix.

\item Designing a dilated encoder-decoder ({\em DiEnDec}) for fusing the full-resolution contextual information of multi-scale Hessian features and inferring the detail-fidelity attention representations in a morphological erosion \& dilation manner, which possesses characteristics of both full-resolution and progressively growing receptive fields.

\item Proposing a learnable distribution alignment cell ({\em DAC}) for adaptively expanding and aligning the attention representation under the prior distribution of referenced features in a statistical manner, which is appropriate for residual attention architectures.
\end{itemize}

\section{Related Work}

Benefiting from the ability of full-resolution deep CNNs to end-to-end non-linear mapping, deep-learning based SR methods have been developed as their excellent performance and real-time processing. Initially, Dong \etal\cite{DongC2016TPAMI} proposed a super-resolution convolutional neural network (SRCNN) to learn a nonlinear LR-to-HR mapping function, which outperforms most conventional example-learning based SR methods. Following this work, deep-learning-based SR methods have achieved excellent performances for decades and been divided into two branches of high-fidelity and lightweight applications:
\subsection{Deep-learning based SR: High-fidelity Type}
Aiming at stacking very deep networks to improve the capability of model to feature representation, Kim \etal proposed two deep networks for accurate SR: VDSR~\cite{KimJ2016CVPR_VDSR} and DRCN~\cite{KimJ2016CVPR_DRCN}, both of which stack more than 20 convolutional layers with global residual learning. And following the residual block in ResNet~\cite{HeK2016CVPR}, local residual learning has been applied in SISR task to stack very deep networks as EDSR~\cite{LimB2017CVPRW} and DRRN~\cite{TaiY2017CVPR}. Nevertheless, the residual learning still exists shortcomings and inefficiency, then MSRN~\cite{LiJ2018ECCV} and OISR~\cite{HeX2019CVPR} were proposed to improve the ability of residual blocks for feature representation.
Moreover, for building stronger relationships crossing information flows in each convolutional layers/blocks, the excellent densely skip connection has been utilized to build very deep model for more accurate super-resolution as MemNet~\cite{TaiY2017ICCV}, SRDenseNet~\cite{TongT2017ICCV}, RDN~\cite{ZhangY2018CVPR} and \etc.
Inspired by the aforementioned trainable attention mechanism, several recent researches were proposed for effectively exploiting the diversity of features, \eg, RCAN~\cite{ZhangY2018ECCV} and CSFM~\cite{HuY2019TCSVT}. Nevertheless, as higher computational complexities with increasing depth of model, these high-fidelity SR methods show difficult implement on mobile applications.
\subsection{Deep-learning based SR: Lightweight Type}
For mobile applications, Dong~\etal\cite{DongC2016ECCV} initially proposed a fast variant of SRCNN, which breaks the bottleneck of computation by straightly using LR image as the input of network and applying deconvolutions into the tail of network for feature upscaling. For similar purpose, Shi~\etal\cite{ShiW2016CVPR} proposed the sub-pixel convolutions to replace the deconvolutions, and achieves comparable performance with relatively less computations. Following them, the lightweight SR gets developed. Representatively, Lai~\etal\cite{LaiW2017CVPR} proposed the LapSRN via progressively reconstructing the sub-band residual HR images at multiple pyramid levels. Besides, several researches were proposed focusing on the lightweight design of network using information diffluence, which splits the intermediate features and processes them separately in manner of connection or convolution, \eg, IDN~\cite{HuiZ2018CVPR} and CARN~\cite{AhnN2018ECCV}.

\section{Proposed Method}
Although stacking very deep networks with small kernel is a feasible way to represent the nonlinear mapping between LR and HR features, it is strongly limited by local consistence and holistically processing each pixel with single convolution in CNNs. Furthermore, in practice, the details and informative textures generally play the dominant roles on visual attention of human vision system, which pays more attentions to the impulse stimulations in scenes and images~\cite{Kastner2000Neuro,Moore2017Psycho}.
Inspired by imitating human visual attentions, we suggest to deal with each components in {\em divide-and-conquer} manner, specifically for SR task, pay more attentions to detail fidelity.

Therefore, in this section, we firstly infer the detail-fidelity attention mechanism and how it works in deep CNNs. Next, we describe an interpretable hessian filtering and its multi-scale variant. Furthermore, we discuss and introduce the dilated encoder-decoder module and distribution alignment cell, which meet the condition of inferring better attention representation of Hessian features.

\subsection{Detail-Fidelity Attention Network}\label{sec:DeFiAN}
It is known that, deep CNNs generally use groups of kernel to convolve with the input features in local receptive fields straightly, \ie, the input features $\{\boldsymbol{x}_i,i=1,2,...,c_{in}\}$ are convolved with the kernel weights $\boldsymbol{w}$ to get the output feature $\{\boldsymbol{y}_j,j=1,2,...,c_{out}\}$ as
\begin{equation}
\boldsymbol{y}_{j}=\sum\limits_{i=1}^{c_{in}}{\boldsymbol{x}_{i}\otimes \boldsymbol{w}_{j,i}}+\boldsymbol{b}_{j}
\end{equation}
where, $c_{in}$ and $c_{out}$ denote the number of input and output channel in a convolution layer.
Specifically, in the receptive field $\mathcal{R}$, for each location $p_*$ on the output feature $\boldsymbol{y}_j$, we have
\begin{equation}
\boldsymbol{y}_{j}({{p}_{*}})=\sum\limits_{i=1}^{c_{in}}{\sum\limits_{p\in \mathcal{R} }{\boldsymbol{x}_{i}({{p}_{*}}+p)\cdot \boldsymbol{w}_{j,i}(p)+{\boldsymbol{b}_{j}}}}
\label{Eq:conventional conv detail}
\end{equation}

Aiming at imitating the human visual attentions in divide-and-conquer manner, it is expected to convolve each patch of local receptive field $\mathcal{R}_*$ with different kernels $\boldsymbol{w}^*$.
However, it is infeasible to apply numerous parametric kernels for all the pixels of image. As a solution, under the assumption of fixed kernel weight $\boldsymbol{w}$, different transformation functions $f^*(\cdot)$ are introduced into Eq.(\ref{Eq:conventional conv detail}):
\begin{equation}
\boldsymbol{y}_{j}({{p}_{*}})=\sum\limits_{i=1}^{c_{in}}{\sum\limits_{p\in \mathcal{R}_* }{\boldsymbol{x}_{i}({{p}_{*}}+p)\cdot f^*(\boldsymbol{w})_{j,i}(p)+{\boldsymbol{b}_{j}}}}
\label{Eq:divide_and_conquer_eq2}
\end{equation}
specific to $\boldsymbol{w}\in \mathbf{R}^{K\times K}$ and $f^*(\boldsymbol{w})\in \boldsymbol{R}^{K\times K}$, each element of a simple linear function $f^*(\boldsymbol{w})$ could be represented as $f^*(\boldsymbol{w})_{j,i}=a^*_{j,i}\cdot\boldsymbol{w}_{j,i}$. Then,
\begin{equation}
\boldsymbol{y}_{j}({{p}_{*}})=\sum\limits_{i=1}^{c_{in}}{\sum\limits_{p\in \mathcal{R}_* }(\boldsymbol{x}_{i}({{p}_{*}}+p)\cdot a^*_{j,i})\cdot \boldsymbol{w}_{j,i}(p)+{\boldsymbol{b}_{j}}}
\label{Eq:divide_and_conquer_eq3}
\end{equation}
particularly, given $a^*_{j,i}$ as a element of matrix $\boldsymbol{a}_{j,i}({{p}_{*}}+p)$, then the divide-and-conquer attention mechanism could be formulated as
\begin{equation}
\boldsymbol{y}_{j}=\sum\limits_{i=1}^{c_{in}}{(\boldsymbol{x}_{i}\cdot\boldsymbol{a}_{j,i})\otimes \boldsymbol{w}_{j,i}}+{\boldsymbol{b}_{j}}
\label{eq:attention mechanism}
\end{equation}

Generally to an integrated system $\boldsymbol{\Phi}$ with divide-and-conquer attention mechanism, the response output $\boldsymbol{y}$ could be formulated as $\boldsymbol{y}=\boldsymbol{\Phi}(\boldsymbol{a}\cdot\boldsymbol{x})$,
where $\boldsymbol{x}$ indicates the input stimulus signal.
Furthermore, specific to SR task, the divide-and-conquer attention matrix $\boldsymbol{a}$ are expected to represent detail fidelity of features $\boldsymbol{x}$.
Therefore, we introduce the detail-fidelity attention module $\{{\textit{DeFiAM}}^{(n)}\}_{n=1}^N$ as the above divide-and-conquer attention system $\boldsymbol{\Phi}$:
\begin{equation}
\begin{aligned}
\boldsymbol{y}&=\textit{DeFiAM}^{(n)}(\boldsymbol{x})\\
&=\boldsymbol{x}+\boldsymbol{\mathcal{F}}^{(n)}(\boldsymbol{x})\cdot\boldsymbol{a}\\
&=\boldsymbol{x}+ \boldsymbol{\mathcal{F}}^{(n)}(\boldsymbol{x})\cdot\boldsymbol{\mathcal{S}}(\boldsymbol{\mathcal{A}}^{(n)}(\boldsymbol{\mathcal{D}}^{(n)}(\boldsymbol{\mathcal{H}}^{(n)}(\boldsymbol{\mathcal{F}}^{(n)}(\boldsymbol{x}))),\boldsymbol{x}))
\end{aligned}
\end{equation}
where $\boldsymbol{\mathcal{F}}^{(n)}(\cdot)$, $\boldsymbol{\mathcal{H}}^{(n)}(\cdot)$, $\boldsymbol{\mathcal{D}}^{(n)}(\cdot)$ and $\boldsymbol{\mathcal{A}}^{(n)}(\cdot)$ denote feature extraction module (\textit{FEM}), Hessian Filtering, dilated encoder-decoder ({\em DiEnDec}) and distribution alignment cell ({\em DAC}) in $n$-th {\em DeFiAM} module, which are parametric and learnable. Besides, the non-linear gate cell $\boldsymbol{\mathcal{S}}(\cdot)$ is implemented using Sigmoid function.

Specifically, $\{\textit{FEM}^{(n)}\}_{n=1}^N$ stacks a chain of $M$ residual channel attention blocks (RCAB) $\{\boldsymbol{\mathcal{B}}^{(n,m)}\}_{m=1}^M$~\cite{ZhangY2018ECCV}, and is formulated as
\begin{equation}
\boldsymbol{\hat{x}}=\boldsymbol{\mathcal{F}}^{(n)}(\boldsymbol{x})=\boldsymbol{\mathcal{B}}^{(n,M)}(\boldsymbol{\mathcal{B}}^{(n,M)}(\cdot\cdot\cdot\boldsymbol{\mathcal{B}}^{(n,M)}(\boldsymbol{x})))
\end{equation}
where ${\mathcal{B}}^{*}(\boldsymbol{x})=\boldsymbol{x}+\boldsymbol{f}_{FE}(\boldsymbol{x})\cdot \boldsymbol{f}_{CA}(\boldsymbol{f}_{FE}(\boldsymbol{x}))$, specifically, each $\boldsymbol{f}_{FE}$ stacks 2 convolution layers with $3\times3$ kernels for feature extraction, and $\boldsymbol{f}_{CA}$ serves for channel attention by stacking global pooling operation, 2 full connection layers and a Sigmoid function successively as~\cite{ZhangY2018ECCV}.

\begin{figure}
	\centering
	\includegraphics[width=0.9\linewidth]{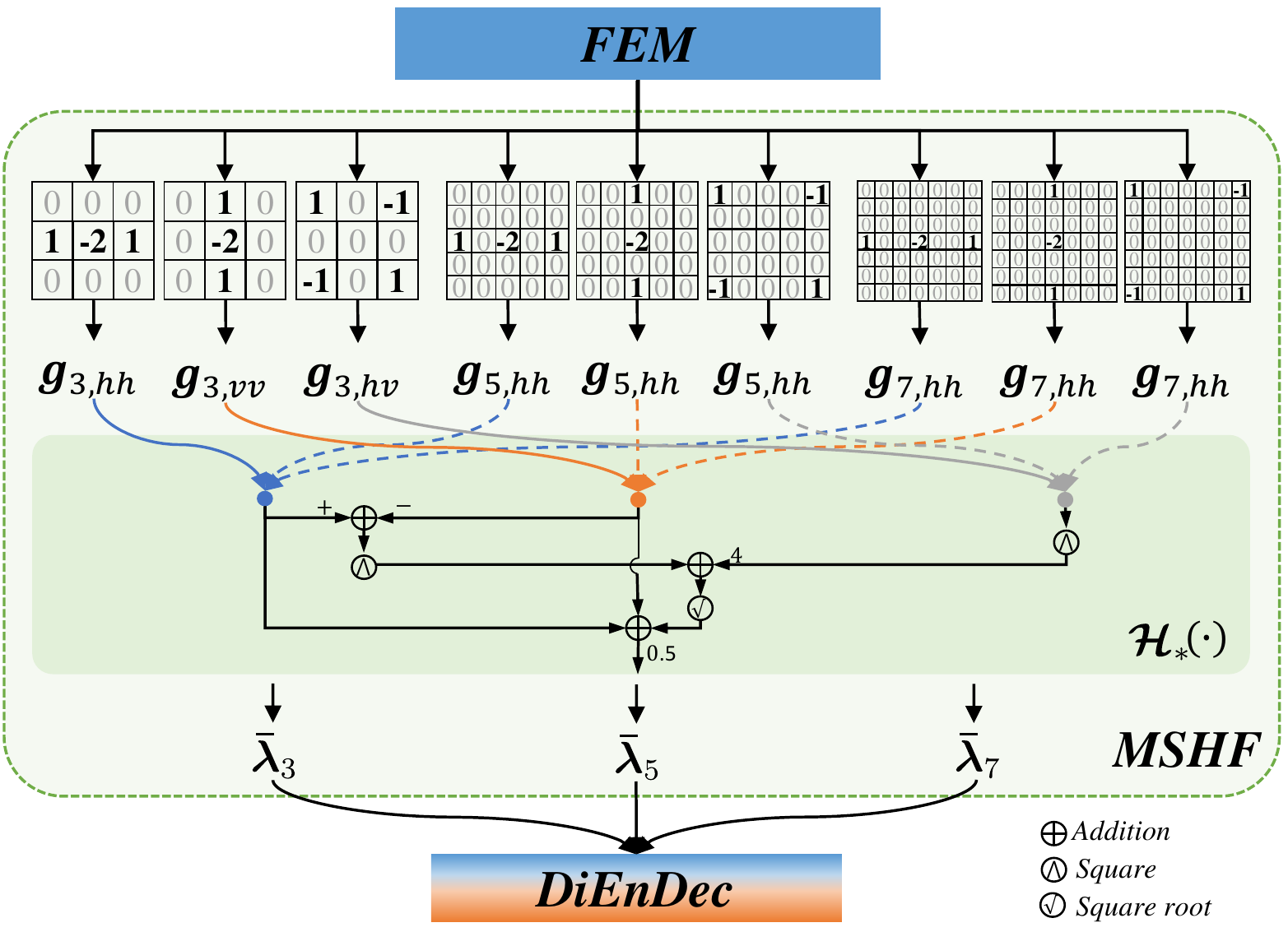}
	\caption{Multi-scale Hessian filtering. Specifically, $\boldsymbol{g}$ denotes the image gradients by applying the gradient-based operator $\boldsymbol{G}(\cdot)$, for example, $\boldsymbol{g}_{3,hh}=\boldsymbol{G}_{3,hh}(\boldsymbol{x})$.}
	\label{fig:MSHF}	
\end{figure}
\subsection{Interpretable Hessian Filtering}\label{sec:Hessian filtering}
As a common criterion for describing the structural characteristics of images at the particular point, Hessian matrix is determined by the second partial derivative of the image in horizontal and vertical directions as
\begin{equation}
\boldsymbol{{H}}(\boldsymbol{x})=\left[ \begin{matrix}
{\boldsymbol{G}_{hh}(\boldsymbol{x})} & {\boldsymbol{G}_{hv}(\boldsymbol{x})} \\
{\boldsymbol{G}_{hv}(\boldsymbol{x})} & {\boldsymbol{G}_{vv}(\boldsymbol{x})} \\
\end{matrix} \right]
\label{Eq:Hessian matrix}
\end{equation}
where the $\boldsymbol{g}_{hh}$=$\boldsymbol{G}_{hh}(\boldsymbol{x})$, $\boldsymbol{g}_{vv}$=$\boldsymbol{G}_{vv}(\boldsymbol{x})$ and $\boldsymbol{g}_{hv}$=$\boldsymbol{G}_{hv}(\boldsymbol{x})$ represent the horizontal, vertical and mixed second partial derivation of $\boldsymbol{x}$ respectively. Particularly, for edge detection~\cite{DengH2007CVPR}, the maximum eigenvalue of this Hessian matrix is an excellent metric to draw the image structural edges with higher intensity and suitable for the residual images, and has positive effects on improving the detail fidelity of image.

\subsubsection{Generic Hessian filtering}
With the development of deep convolutional neural networks, almost gradient-based feature detection operators are attainable by using the convolutions with fixed weights on the efficient GPUs. From the Eq.(\ref{Eq:Hessian matrix}), the Hessian matrix of $\boldsymbol{x}$ depends on its second order derivations which are available by applying some specific filters:
\begin{equation}
\boldsymbol{G}_{hh}=\left[ \begin{matrix}
\text{0} & \text{0} & \text{0} \\
\text{1} & \text{-2} & \text{1} \\
\text{0} & \text{0} & \text{0} \\
\end{matrix} \right], \boldsymbol{G}_{vv}=\boldsymbol{G}_{hh}^\text{T},\boldsymbol{G}_{hv}=\left[ \begin{matrix}
1 & \text{0} & -1 \\
\text{0} & \text{0} & 0 \\
-1 & \text{0} & \text{1} \\
\end{matrix} \right]
\label{Eq:Specific filters}
\end{equation}

Furthermore, the maximum eigenvalue of Hessian matrix is necessary for representing the specific detail-fidelity features, which generally is calculated by applying SVD algorithm or eigenvalue decomposition and is hard to calculate with the acceleration of GPUs.
Then, since the Hessian matrix $\boldsymbol{{H}}(\boldsymbol{x})$ is real symmetric, we then utilize the linear algebra to infer the relationships between eigenvalues (${\boldsymbol{\lambda}_1}>{\boldsymbol{\lambda}_2}$), trace $tr(\boldsymbol{{H}}(\boldsymbol{x}))$ and determinant $\left| \boldsymbol{{H}}(\boldsymbol{x}) \right|$ of such real symmetric matrix:
\begin{equation}
tr(\boldsymbol{{H}}(\boldsymbol{x})) = {\boldsymbol{G}_{hh}(\boldsymbol{x})} + {\boldsymbol{G}_{vv}(\boldsymbol{x})}={{\boldsymbol{\lambda} }_{1}}+{{\boldsymbol{\lambda} }_{2}}
\label{Eq:Trace}
\end{equation}
\begin{equation}
\left| \boldsymbol{{H}}(\boldsymbol{x}) \right|= {\boldsymbol{G}_{hh}(\boldsymbol{x})} {\boldsymbol{G}_{vv}(\boldsymbol{x})} - {\boldsymbol{G}^2_{hv}(\boldsymbol{x})} ={{\boldsymbol{\lambda} }_{1}}{{\boldsymbol{\lambda} }_{2}}
\label{Eq:Det}
\end{equation}

By solving this linear formulation and being free of calculating the intricate inverse of matrix, the maximum eigenvalue $\boldsymbol{\lambda}$ ($\boldsymbol{\lambda}=\boldsymbol{\lambda}_1$) of $\boldsymbol{{H}}(\boldsymbol{x})$ becomes a concise algebraic combination of the image gradients as
\begin{equation}
\begin{aligned}
\boldsymbol{\lambda}&=\boldsymbol{\mathcal{H}}(\boldsymbol{x})\\
&=({\boldsymbol{G}_{hh}(\boldsymbol{x})}+{\boldsymbol{G}_{vv}(\boldsymbol{x})})/2+\\
&\text{ }\text{ }\text{ }\text{ } \left(\sqrt{{{({\boldsymbol{G}_{hh}(\boldsymbol{x})}-{\boldsymbol{G}_{vv}(\boldsymbol{x})})}^{2}}+4\boldsymbol{G}_{hv}^{2}(\boldsymbol{x})}\right)/2
\end{aligned}
\label{Eq:Eigenvalues}
\end{equation}
since only the the pixel-wise gradients are required, it is accessible to utilize the specifical convolutions as Eq.(\ref{Eq:Specific filters}) to accelerate this {\em Hessian Filtering (HF)} procedure $\boldsymbol{\mathcal{H}}(\cdot)$ and then obtain the definitive $\lambda\in\boldsymbol{\lambda}$ for all the pixel-wise elements $x\in\boldsymbol{x}$.

More generically, $\boldsymbol{\lambda}$ is an algebra combination of the second order gradients of $\boldsymbol{x}$, and
\begin{equation}
{\boldsymbol{\lambda}}\propto \sum\limits_{i=-1,0,1}{\nabla \boldsymbol{x}^{i}}
\label{Eq: lambda}
\end{equation}
where $\nabla$ represents the gradient extractor via Eq.(\ref{Eq:Specific filters}).

\begin{figure*}
	\begin{minipage}[b]{1.0\linewidth}
		\subfloat[parameterization and $ARF$ of {\em DiEnDec}]{
			\centering
			\includegraphics[width=0.6\linewidth]{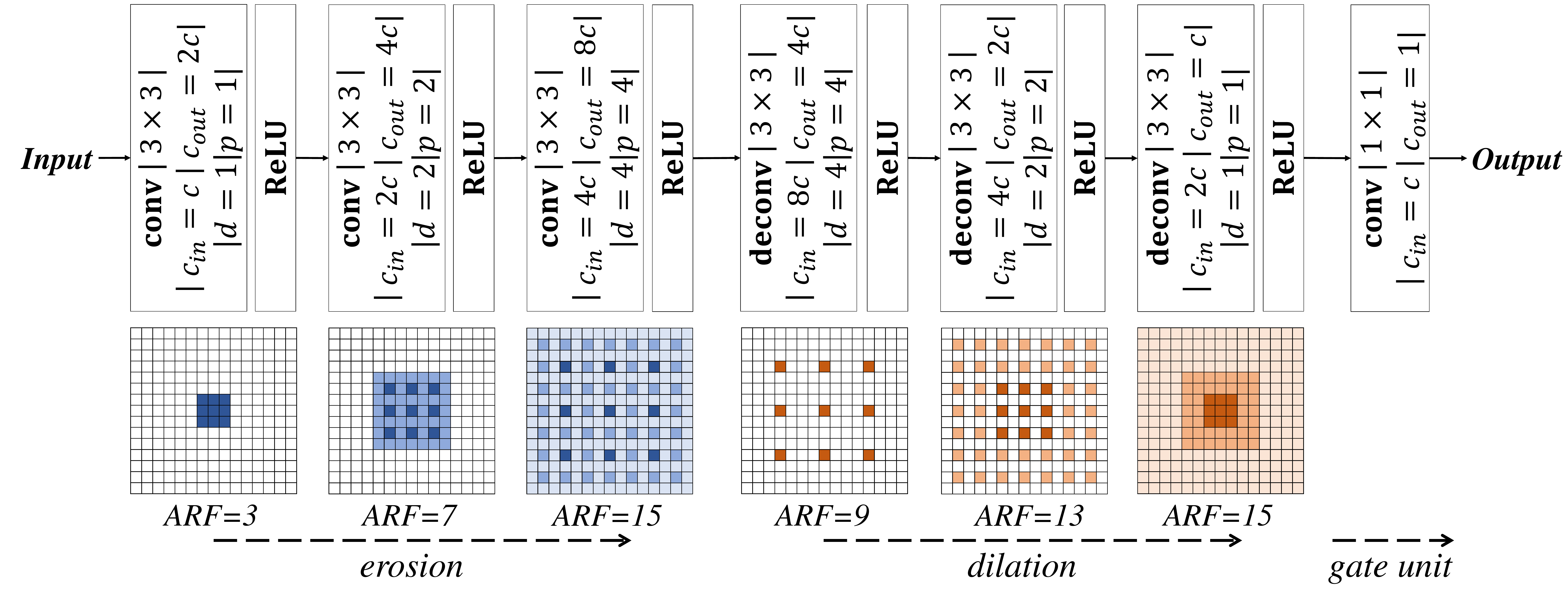}
		}
        \subfloat[structures of encoder-decoder]{
		\centering
	       \includegraphics[width=0.35\linewidth]{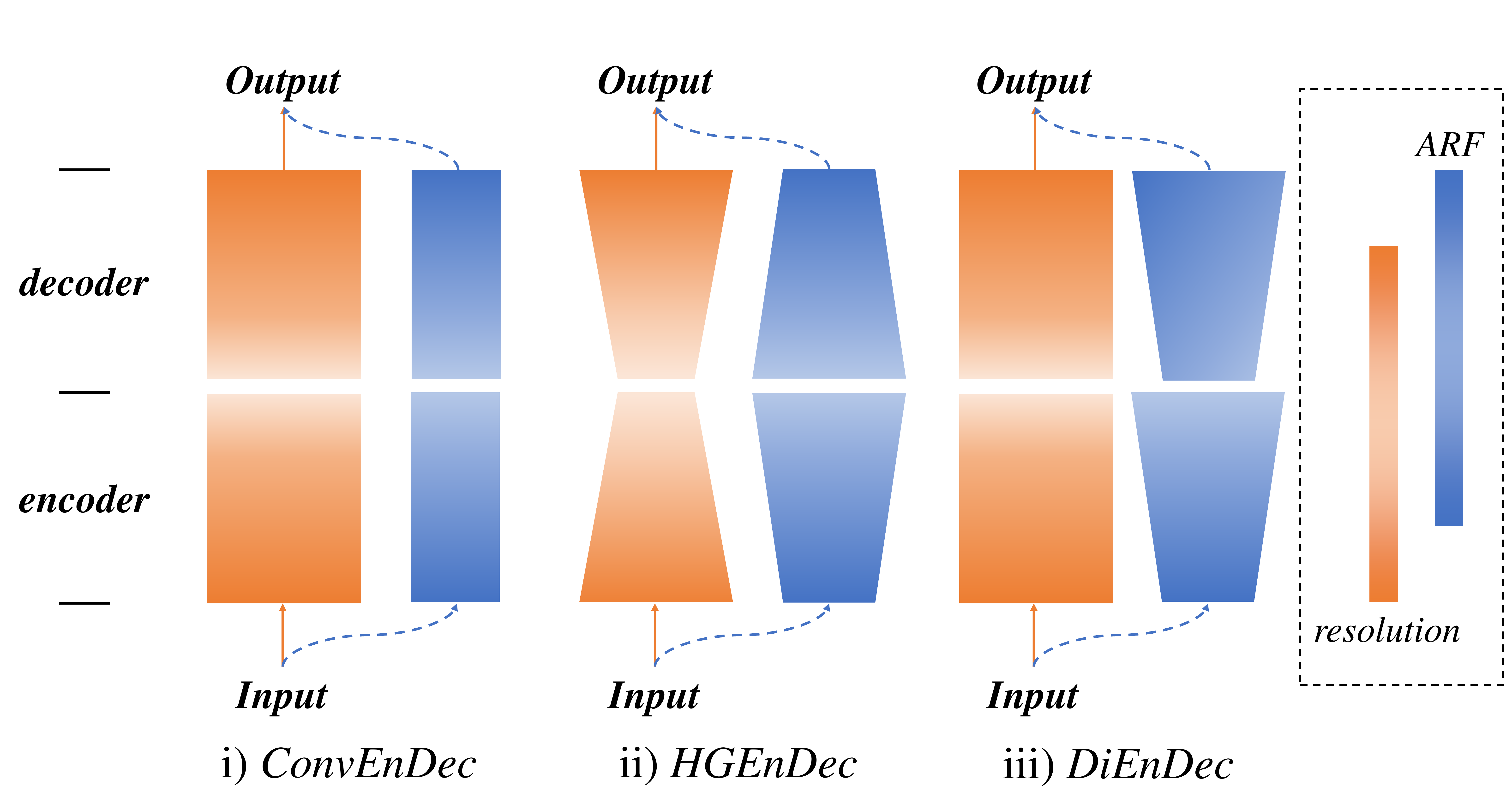}
		}
	\end{minipage}
	\caption{Illustration on {\em DiEnDec}: (a) Parameterization and accumulative receptive field ($ARF$) of {\em DiEnDec} ($d$ and $p$ represent the dilation and padding of convolution, respectively; (b) Structures of encoder-decoder: i) convolutional encoder-decoder({\em ConvEnDec})~\cite{MaoX2016NIPS}; ii) hourglass encoder-decoder ({\em HGEnDec})~\cite{Newell2016ECCV}; iii) the proposed {\em DiEnDec}, specifically, the orange and blue path indicate the view of resolution and receptive field, obviously {\em DiEnDec} equips characteristics of both full-resolution and progressively growing receptive fields.}
	\label{fig:DiEnDec}
\end{figure*}
\subsubsection{Multi-scale Hessian filtering}\label{sec:MSHF}
Furthermore, in deep residual networks for low-level vision, the residual features are becoming finer and thinner with the increasing depth~\cite{HuangY2019ICME}. That is, the features learned in shallower architecture would represent coarser residues such as geometric structures, as the architecture becomes deeper, more details such as textures and even noises could be generated.

Therefore, to extract various detail-fidelity components of image/feature, we introduce a scaled variant of the above generic Hessian filtering, which is named {\em scaled Hessian Filtering} and denoted as $\boldsymbol{\mathcal{H}}_{ker}(\cdot)$, where $ker$ represents the kernel size of the specific filters in Eq.(\ref{Eq:Specific filters}) by padding the zero-vectors. For example,\\
(1) for $\boldsymbol{\mathcal{H}}_{3}(\cdot)$ filtering (generic Hessian filtering), the specific gradient extractors are
\begin{equation}
\begin{aligned}
&\boldsymbol{G}_{3, hh}=\left[\text{1}, \text{-2}, \text{1} \right]\\
&\boldsymbol{G}_{3, vv}=\left[\text{1}, \text{-2}, \text{1} \right]^\text{T}\\
&\boldsymbol{G}_{3, hv}=\left[\text{1}, \text{0}, \text{-1} \right]^\text{T}\left[\text{1}, \text{0}, \text{-1} \right]\\
\end{aligned}
\end{equation}
then its corresponding maximum eigenvalue of Hessian could be inferred as
\begin{equation}
{\boldsymbol{\lambda}_3}\propto \sum\limits_{i=-1,0,1}{\nabla \boldsymbol{x}^{i}}
\label{Eq: lambda_3}
\end{equation}
\\
(2) for $\boldsymbol{\mathcal{H}}_{5}(\cdot)$ filtering, the specific gradient extractors are
\begin{equation}
\begin{aligned}
&\boldsymbol{G}_{5, hh}=\left[\text{1}, \text{0}, \text{-2}, \text{0}, \text{1} \right]\\
&\boldsymbol{G}_{5, vv}=\left[\text{1}, \text{0}, \text{-2}, \text{0}, \text{1} \right]^\text{T}\\
&\boldsymbol{G}_{5, hv}=\left[\text{1}, \mathbf{0}, \text{-1} \right]^\text{T}\left[\text{1}, \mathbf{0}, \text{-1} \right]\\
\end{aligned}
\end{equation}
where $\{\mathbf{0}\}$ represents a 3-dimension zero vector.
Then its corresponding maximum eigenvalue of Hessian could be inferred as
\begin{equation}
{\boldsymbol{\lambda}_5}\propto \sum\limits_{i=-2,0,2}{\nabla \boldsymbol{x}^{i}}
\label{Eq: lambda_5}
\end{equation}

In this way, a smaller scaled Hessian filter (\eg, $\boldsymbol{\mathcal{H}}_{3}(\cdot)$) pays more attentions to infer the lower-order maximum eigenvalue of Hessian and exploits the gradients of closer neighbored pixels to generate finer information, \eg, the edges and textures. Instead, a larger one (\eg, $\boldsymbol{\mathcal{H}}_{5}(\cdot)$) pays more attentions to infer higher-order maximum eigenvalue of Hessian and generate coarser information, \eg, the structures.
Therefore, we utilize the multi-scale Hessian filters to infer various and informative details on each input feature $\boldsymbol{x}$. As shown in Fig.~\ref{fig:MSHF}, three scald Hessian filters $\boldsymbol{\mathcal{H}}_{3}(\cdot)$, $\boldsymbol{\mathcal{H}}_{5}(\cdot)$, and $\boldsymbol{\mathcal{H}}_{7}(\cdot)$ are exploited to implementing the {\em Multi-Scale Hessian Filtering (MSHF)} procedure to infer various interpretable detail-fidelity features, \ie, maximum eigenvalue of multi-scale Hessian matrix.

Particularly, since the size of the eigenvalue $\boldsymbol{\lambda}_{ker}$ is same as the corresponding inputs $\boldsymbol{x}$, and under the assumption of Eq.(\ref{Eq:divide_and_conquer_eq3}) that the size of detail-fidelity attention $\boldsymbol{a}$ should also be same as $\boldsymbol{x}$. Therefore, a fusion unit is necessary to infer the detail-fidelity attentions from the multi-scale Hessian eigenvalues, and formulated as
\begin{equation}
{\boldsymbol{a}} = \boldsymbol{f}_{FU}([\boldsymbol{\bar{\lambda}}_{3}, \boldsymbol{\bar{\lambda}}_{5}, \boldsymbol{\bar{\lambda}}_{7}])
\label{eq:MSHF_fusion}
\end{equation}
specifically, the multi-scale Hessian eigenvalues are averaged in channel-wise as $\boldsymbol{\bar{\lambda}}_{*}$ to alleviate computational complexity. Particularly, as Eqs.(\ref{Eq: lambda_3}) and (\ref{Eq: lambda_5}), the maximum eigenvalues of multi-scale Hessian matrix are proportional to the multi-order gradients of inputs, which only represent the thin residues but are incapable of illustrating attention maps. Therefore, aiming at both inferring the details of feature and capable of attention representation, the fusion unit $\boldsymbol{f}_{FU}$ should meet several conditions:\\
(1) {\em Full-resolution manner} for the assumption of output $\boldsymbol{a}$ and input $\boldsymbol{x}$ with the same size;\\
(2) {\em Channel extension manner} for the dot product and summation operations of Eq.(\ref{Eq:divide_and_conquer_eq3});\\
(3) {\em Erosion \& dilation manner} for attention representation.

Under these predetermined conditions, we design a dilated encoder-decoder ({\em DiEnDec}) with full-resolution and erosion \& dilation manners in morphology, and a distribution alignment cell ({\em DAC}) for channel extension manner in statistic.

\subsection{Dilated Encoder-Decoder}\label{sec:DiEnDec}
Generally, an end-to-end architecture for low-level vision concerns more about the pixel intensities in full-resolution and aims at producing the expected details in a coarse-to-fine fashion which is contrary to the fashion in semantic segmentation or image classification. Therefore, full-convolution architecture is generally exploited for low-level vision tasks and meets the requested full-resolution manner.

Moreover, in order to learn some specific contextual information, namely implementation of erosion \& dilation manner, encoder-decoder is designed to transform the input features into specific contextual styles. To utilize the global contextual information of images, hourglass-like encoder-decoder ({\em HGEnDec})~\cite{Balle2017ICLR,Newell2016ECCV} are designed by combination of general convolutions and pooling operators. However, in the case of super-resolution, pooling is inappropriate to retain the full resolution. Then, Mao~\etal\cite{MaoX2016NIPS} present a full-resolution convolutional encoder-decoder ({\em ConvEnDec}) for image restoration, in this way, the encoder cascaded by a group of $3\times3$ convolutions suppresses the noises and the decoder with deconvolution reconstructs the denoised features. However, this method only considers the information of small $3\times3$ local receptive field for each pixel and lacks contextual information.
Therefore, {\em a full-convolution and context-aware architecture is necessary}.

Beyond the limitation of general convolution, the dilated convolution~\cite{YuF2016ICLR} considers a wider range of neighbors for each pixel with dilation in convolution and then aggregates more contextual information in full-resolution fashion. As Fig.~\ref{fig:DiEnDec}(b) shows and differ from the existing encoder-decoders, we explore the ability of both full-resolution and extracting contextual information by introducing the dilated convolution, named Dilated Encoder-Decoder ({\em DiEnDec}) and parameterized as Fig.~\ref{fig:DiEnDec}(a).
In each {\em DiEnDec}, an exponential expansion of receptive field with no gridding effects~\cite{WangP2018WACV} is available by using the exponential increasing dilation of convolution. For the $k\times k$ kernel of dilated convolution with dilation of ${(k-1)}^i$ in the encoder with $d$ effective layers, the accumulative receptive field ({\small $ARF$}) could be formulated as
\begin{equation}
ARF=1+\sum\limits_{i=1}^{M}{{{(k-1)}^{i}}}
\label{Eq:Receptive field}
\end{equation}

By reverse thinking, we use the dilated deconvolution to reproducing the information by decoding the contextual features. For better viewing, Fig.~\ref{fig:DiEnDec}(a) shows the available {\small $ARF$} in each stage of encoding and decoding. Morphologically, by using these dilated convolutions, the encoder not only suppresses the noisy features and also obtain the neighboring relationships in larger receptive fields for feature erosion, then the decoder reconstruct these obscure eroded features into more representative dilated styles.

Specifically, under the assumption of $\boldsymbol{x} \in \mathbf{R}^{c\times w\times h}$ and $\boldsymbol{\mathcal F}(\boldsymbol{x}) \in \mathbf{R}^{c\times w\times h}$, where $c$, $w$ and $h$ denote the channel size, width and height of features. As in Fig.~\ref{fig:DiEnDec}(a), aiming at inferring the representative attention features with lower computational complexity, the output of {\em DiEnDec} is $\boldsymbol{v}=\boldsymbol{\mathcal{D}}([\boldsymbol{\bar{\lambda}}_{3}, \boldsymbol{\bar{\lambda}}_{5}, \boldsymbol{\bar{\lambda}}_{7}]) \in \mathbf{R}^{1\times w\times h}$ with context-aware attention representation.

\subsection{Distribution Alignment Cell}~\label{sec:DAC}
To meet the dot production and summation operations of Eq.~(\ref{Eq:divide_and_conquer_eq3}) for $\boldsymbol{x}\in\mathbf{R}^{c\times w\times h}$, it is necessary to expand the detail-fidelity attentions $\boldsymbol{a}\in \mathbf{R}^{c\times w\times h}$ from the above attention representation $\boldsymbol{v}\in\mathbf{R}^{1\times w\times h}$. Specifically, as Eq.(\ref{Eq:divide_and_conquer_eq3}), in each {\em DeFiAM}, the final output feature $\boldsymbol{y}$ is a linear representation of input features $\boldsymbol{x}$, residual features $\boldsymbol{\mathcal{F}}(\boldsymbol{x})$ and attention representation $\boldsymbol{a}$. Then it is necessary to align the attention representation and features for optimal convergency, and for this issue, the channel attention was proposed for feature recalibration in channel-wise~\cite{ZhangY2018ECCV,HuJ2018CVPR}, but with the superimposed effect that the observed features are enhanced with the channel-wise representation scalars of referenced features, and is generally helpless for alignment.

\begin{figure}
	\centering
	\includegraphics[width=1\linewidth]{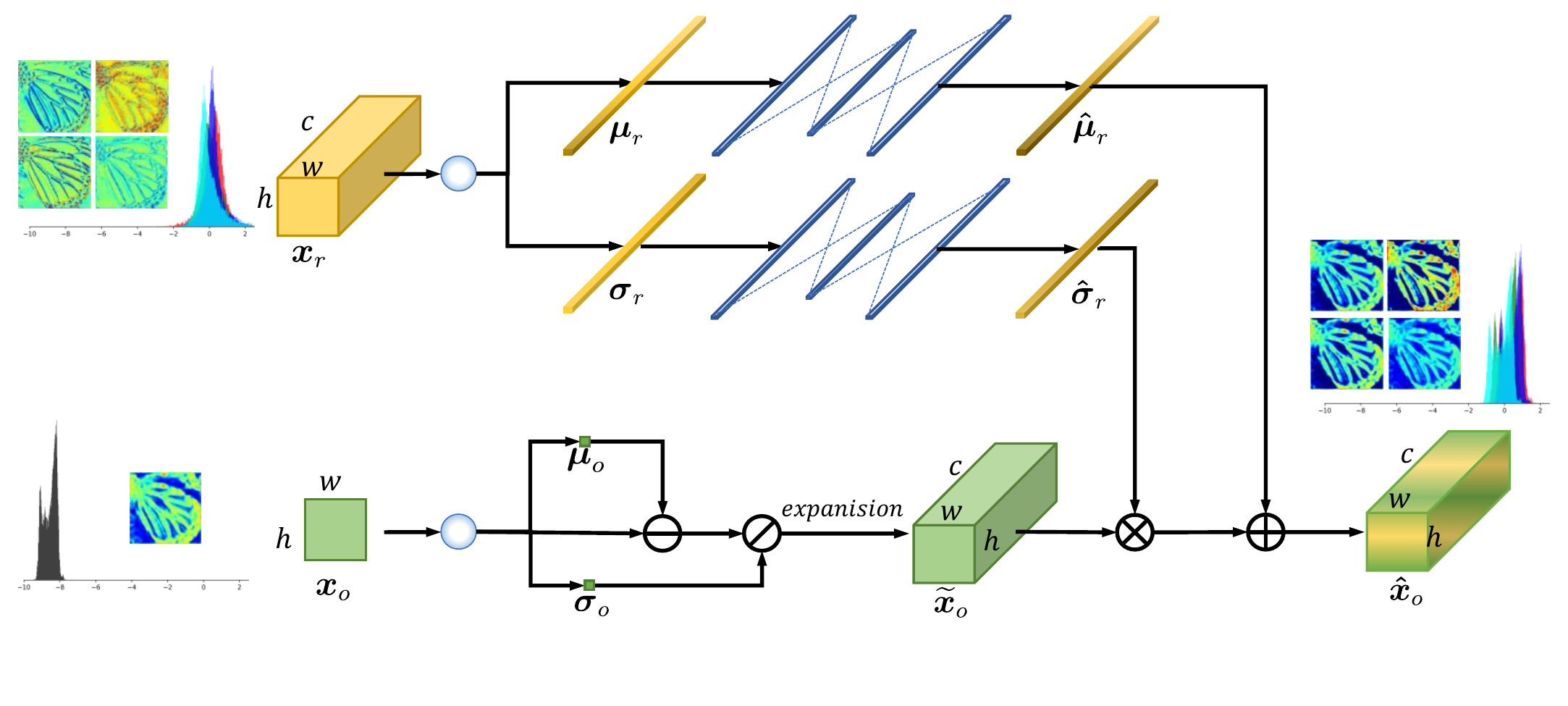}
	\caption{Distribution alignment cell ({\em DAC}) for aligning the observed feature on the prior distribution of referenced features. By using {\em DAC}, the expanded observed features $\boldsymbol{\tilde{x}}$ reserve its distribution in spatial-wise, and also meet similar distribution as the referenced features in channel-wise.}
	\label{fig:DAC}
\end{figure}

We then propose a distribution alignment cell ({\em DAC}) to expand the attention representation $\boldsymbol{v}$ under the prior distribution of input features $\boldsymbol{x}$, as illustrated in Fig.~\ref{fig:DAC}. Particularly, for a better understand of this subsection, we use $\boldsymbol{x}_o$ (observed feature) and $\boldsymbol{x}_r$ (referenced features) replace $\boldsymbol{v}$ and $\boldsymbol{x}$, respectively.
Under the assumption that both of the observed and referenced features meet some prior distributions, \eg, assume $\boldsymbol{x}_o\sim \mathbb{N}(\boldsymbol{\mu}_o, \boldsymbol{\sigma}^2_o)$ and $\boldsymbol{x}_r\sim \mathbb{N}(\boldsymbol{\mu}_r, \boldsymbol{\sigma}^2_r)$, {\em DAC} firstly normalize the observed feature by
\begin{equation}
\boldsymbol{\tilde{x}}_o=\frac{\boldsymbol{x}_o-\boldsymbol{\mu}_o}{\sqrt{\boldsymbol{\sigma}^2_o+\epsilon }}
\label{eq:normalization}
\end{equation}
and expand it into $\boldsymbol{\tilde{x}}\in\mathbf{R}^{c\times w\times h}$ via channel-wise replication.

Besides, {\em DAC} also parameterize the distribution of referenced features using nonlinear neural networks as
\begin{equation}
\begin{aligned}
&\boldsymbol{\hat{\mu}}_r=\boldsymbol{f}_{FC}(\boldsymbol{f}_{ReLU}(\boldsymbol{f}_{FC}(\boldsymbol{\mu}_r)))\\
&\boldsymbol{\hat{\sigma}}_r=\boldsymbol{f}_{FC}(\boldsymbol{f}_{ReLU}(\boldsymbol{f}_{FC}(\boldsymbol{\sigma}_r)))
\end{aligned}
\label{eq:paramerization_DAC}
\end{equation}
where, $\boldsymbol{f}_{FC}$ and $\boldsymbol{f}_{ReLU}$ represent the full connection layer and Rectified Linear Unit (ReLU), respectively.

Then since the normalized observed feature meets the standard normal distribution, it is feasible to align them with the parameterized priors, as
\begin{equation}
\boldsymbol{\hat{x}}_o=\boldsymbol{\tilde{x}}_o\cdot \boldsymbol{\hat{\sigma}}_r+\boldsymbol{\hat{\mu}}_r
\label{eq:alignment}
\end{equation}
then, the aligned observed features meet similar distribution as the referenced features, namely, $\boldsymbol{\hat x}_o\sim \mathbb{N}(\boldsymbol{\hat\mu}_r, \boldsymbol{\hat\sigma}^2_r)$.

To be emphasized, since the {\em DAC} procedure only works in channel-wise, the spatial-wise distribution of $\boldsymbol{x}_o$ has been also retained for attention representation.

\section{Experiment}
In this section, we first provide the benchmarks (including datasets and evaluation criterion) and implementation details (including hyper-parameters and optimization). Then we compare our {\em DeFiAN} model with other state-of-the-art methods on several benchmark datasets. Next we study the contributions of different components in the proposed {\em DeFiAN} by experimental demonstrations and finally illustrate discussion on limitation.

\subsection{Benchmarks}
\subsubsection{Datasets}
In testing phase, we implement our experiments on several benchmark datasets for evaluation, including Set5~\cite{BevilacC2012BMVC}, Set14~\cite{ZeydeR2010}, BSD100~\cite{ArbelaezP2011TPAMI}, Urban100~\cite{HuangJB2015CVPR} and Manga109~\cite{LaiW2017CVPR} with general, structural and cartoon scenes.
In training phase, 800 high-quality 2K-resolution images from DIV2K~\cite{TimofteR2017CVPRW} dataset are considered as high-resolution groundtruth for training the proposed models, and down-sampled using Bicubic algorithm to generate the corresponding low-resolution inputs. In detail, we use the $48\times48$ RGB patches from the low-resolution training set as input and the corresponding $48s\times48s$ high-resolution patches as groundtruth for $\times s$ upscaling, and augment these LR-HR pairs with random horizontal and vertical flips and 90 rotations. Particularly, all the LR and HR images are pre-processed by subtracting the mean RGB value of the training sets.

\subsubsection{Evaluation}
For evaluating the SR performance, we apply two common full-reference image quality assessment criteria for evaluating discrepancies: Peak Signal-to-Noise Ratio (PSNR) and Structural SIMilarity (SSIM). Following the convention of super-resolution, luminance channel is selected for full-reference image quality assessment because the intensity of image is more sensitive to human-vision than the chroma. Moreover, we use two criteria to represent the computational complexities of methods, where \#Params denotes the number of parameters in whole network, \#FLOPs indicates the number of operations by Multi-Adds which is the number of composite multiply-accumulate operations for processing a $480\times360\times3$ RGB image, which indicate the space complexity and time complexity respectively.

\begin{table*}
	\centering
    \captionsetup{justification=centering}
	\caption{\textsc{Quantitative comparisons of the proposed $\textit{DeFiAN}_L$ with the state-of-the-art \\high-fidelity methods on benchmark datasets for $\times2$, $\times3$ and $\times4$ upscaling.}}
	\begin{tabular}{{c c c c c c c c c c c c c c}}
	\hline
    \hline
    \multirow{2}{*}{Scale}&\multirow{2}{*}{Method}&\#Params&\#FLOPs
    &\multicolumn{2}{c}{Set5}&\multicolumn{2}{c}{Set14}&\multicolumn{2}{c}{BSD100}
    &\multicolumn{2}{c}{Urban100}&\multicolumn{2}{c}{Manga109}\\
    &&(K)&(G)&PSNR&SSIM&PSNR&SSIM&PSNR&SSIM&PSNR&SSIM&PSNR&SSIM\\
	\hline
	\multirow{10}{*}{$\times2$}
    &Bicubic&-&-
    &33.64&0.9292&30.22&0.8683&29.55&0.8425&26.87&0.8397&30.80&0.9339\\
	&MemNet~\cite{TaiY2017ICCV}&2905.5&503.5
    &37.78&0.9597&33.28&0.9142&32.08&0.8978&31.31&0.9195&38.02&0.9755\\
	&EDSR~\cite{LimB2017CVPRW}&40711.7&1760.0
    &38.11&0.9600&33.82&0.9189&32.33&0.9011&32.94&0.9351&39.10&0.9773\\
	&D-DBPN~\cite{Haris2018CVPR}&5953.5&270.0
    &38.05&0.9599&33.70&0.9188&32.25&0.9001&32.51&0.9316&38.81&0.9766\\
	&RDN~\cite{ZhangY2018CVPR}&5616.3&243.0
    &38.16&0.9603&33.88&0.9199&32.31&0.9009&32.89&0.9353&39.09&0.9771\\
	&MSRN~\cite{LiJ2018ECCV}&5930.1&256.5
    &37.98&0.9597&33.56&0.9171&32.20&0.8994&32.29&0.9296&38.55&0.9762\\
	&RCAN~\cite{ZhangY2018ECCV}&15444.7&663.5
    &\underline{38.18}&0.9604&\underline{34.00}&\underline{0.9203}&\underline{32.37}&\underline{0.9016}&
\underline{33.14}&\underline{0.9364}&\underline{39.34}&\underline{0.9777}\\
	&CSFM~\cite{HuY2019TCSVT}&12071.9&519.9
    &38.17&\underline{0.9605}&33.94&0.9200&32.34&0.9013&33.08&0.9358&39.30&0.9775\\
	&OISR-RK3~\cite{HeX2019CVPR}&44270.0&1812.1
    &38.13&0.9602&33.81&0.9194&32.34&0.9011&33.00&0.9357&39.13&0.9773\\
	&$\textit{DeFiAN}_L$(Ours)&15186.1&651.4
    &{\bf38.33}&{\bf0.9618}&{\bf34.28}&{\bf0.9231}&{\bf32.43}&{\bf0.9029}&{\bf33.39}&{\bf0.9390}&{\bf39.78}&{\bf0.9797}\\
	\hline
	\multirow{9}{*}{$\times3$}
    &Bicubic&-&-
    &30.40&0.8686&27.54&0.7741&27.21&0.7389&24.46&0.7349&26.95&0.8565\\
	&MemNet~\cite{TaiY2017ICCV}&2905.5&503.5
    &34.12&0.9254&30.03&0.8356&28.98&0.8013&27.60&0.8393&32.79&0.9389\\
	&EDSR~\cite{LimB2017CVPRW}&43660.8&839.6
    &34.64&0.9283&30.42&0.8447&29.26&0.8089&28.80&0.8652&34.17&0.9476\\
	&RDN~\cite{ZhangY2018CVPR}&5800.6&111.7
    &34.67&0.9285&30.46&0.8451&29.24&0.8084&28.80&0.8653&34.09&0.9474\\
	&MSRN~\cite{LiJ2018ECCV}&6115.0&117.7
    &34.45&0.9265&30.29&0.8418&29.13&0.8054&28.30&0.8553&33.53&0.9441\\
	&RCAN~\cite{ZhangY2018ECCV}&15592.4.8&290.8
    &\underline{34.72}&0.9289&\underline{30.52}&0.8458&\underline{29.29}&\underline{0.8099}&\underline{29.00}
    &\underline{0.8679}&34.39&0.9488\\
	&CSFM~\cite{HuY2019TCSVT}&12256.5&234.8
    &\underline{34.72}&\underline{0.9290}&\underline{30.52}&\underline{0.8459}&\underline{29.29}&0.8096&
    28.97&0.8673&\underline{34.48}&\underline{0.9492}\\
	&OISR-RK3~\cite{HeX2019CVPR}&44877.2&862.7
    &34.69&0.9287&30.46&0.8454&29.28&0.8096&28.94&0.8674&34.28&0.9483\\
	&$\textit{DeFiAN}_L$(Ours)&15370.4&293.2
    &{\bf34.81}&{\bf0.9305}&{\bf30.75}&{\bf0.8493}&{\bf29.37}&{\bf0.8120}&{\bf29.28}&{\bf0.8733}&{\bf34.84}&{\bf0.9515}\\
	\hline
	\multirow{10}{*}{$\times4$}
    &Bicubic&-&-
    &28.42&0.8101&25.99&0.7023&25.96&0.6672&23.14&0.6573&24.89&0.7866\\
	&MemNet~\cite{TaiY2017ICCV}&2905.5&503.5
    &31.74&0.8900&28.23&0.7731&27.43&0.7291&25.54&0.7666&29.64&0.8968\\
	&EDSR~\cite{LimB2017CVPRW}&43071.0&542.9
    &32.46&0.8976&28.71&0.7857&27.72&0.7414&26.64&0.8029&31.02&0.9148\\
	&RDN~\cite{ZhangY2018CVPR}&5763.7&67.4
    &32.45&0.8979&28.70&0.7849&27.71&0.7410&26.61&0.8020&30.98&0.9141\\
	&D-DBPN~\cite{Haris2018CVPR}&10426.4&128.5
    &32.40&0.8966&28.66&0.7839&27.67&0.7385&26.38&0.7938&30.89&0.9127\\
	&MSRN~\cite{LiJ2018ECCV}&6077.5&70.7
    &32.22&0.8947&28.53&0.7811&27.60&0.7371&26.20&0.7898&30.55&0.9093\\
	&RCAN~\cite{ZhangY2018ECCV}&15197.0&168.1
    &\underline{32.62}&\underline{0.8992}&28.75&0.7862&27.74&0.7424&26.74&0.8058&31.18&0.9160\\
	&CSFM~\cite{HuY2019TCSVT}&12219.6&136.6
    &32.57&0.8988&\underline{28.77}&\underline{0.7864}&\underline{27.75}&\underline{0.7424}&26.77&0.8057&
    \underline{31.30}&\underline{0.9172}\\
	&OISR-RK3~\cite{HeX2019CVPR}&44287.1&555.9
    &32.51&0.8984&28.76&0.7863&\underline{27.75}&0.7423&\underline{26.78}&\underline{0.8065}&31.24&0.9160\\
	&$\textit{DeFiAN}_L$(Ours)&15333.5&169.5
    &\bf{32.67}&\bf{0.9009}&\bf{28.99}&\bf{0.7906}&\bf{27.84}&\bf{0.7448}&\bf{27.05}&\bf{0.8134}&\bf{31.67}&\bf{0.9211}\\
    \hline
    \hline
	\end{tabular}
	\label{tab:full-reference_comparison_highfidelity}
\end{table*}
\begin{figure*}
    \captionsetup[subfloat]{justification=centering}
	\subfloat[Ground truth HR \protect\\ "{\em img078}"]{
		\centering
		\includegraphics[width=0.208\linewidth]{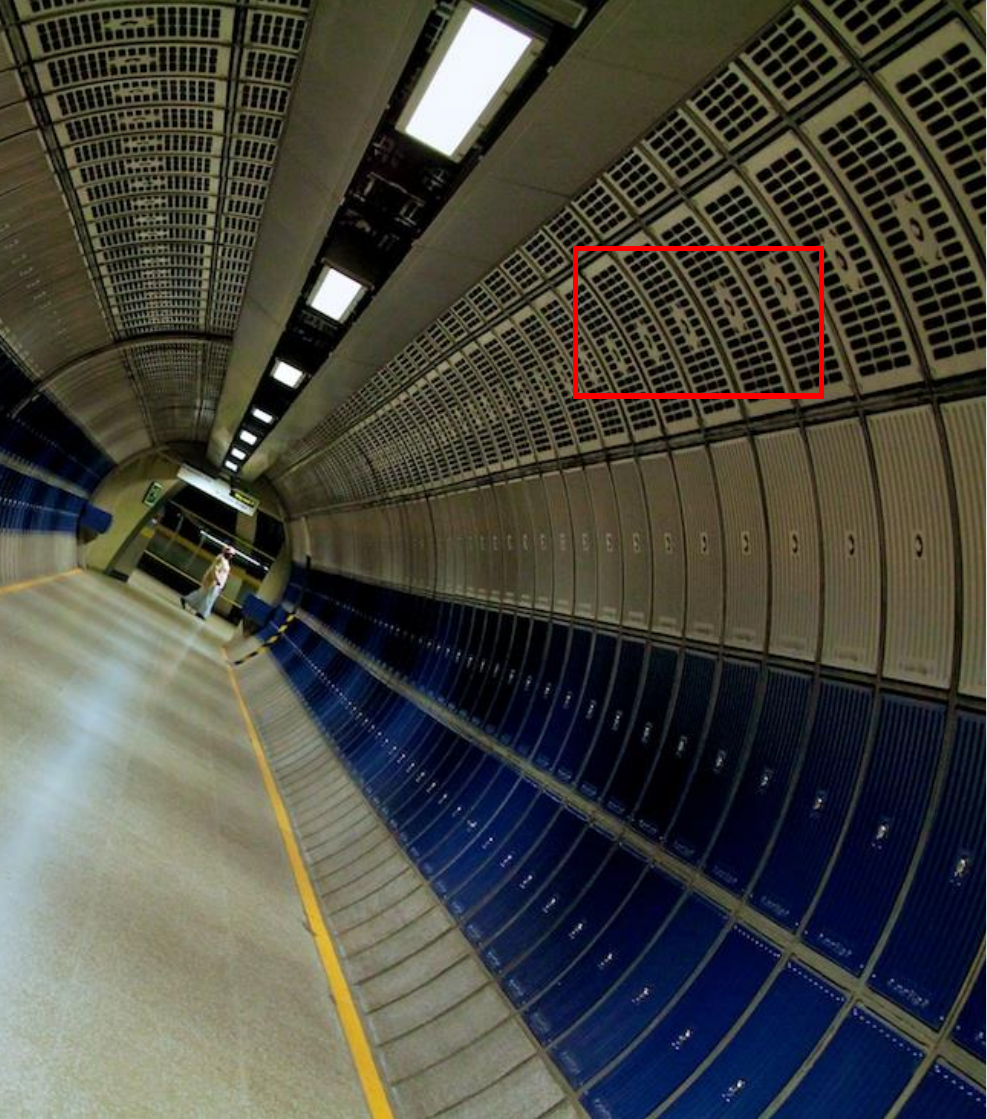}
	}
	\begin{minipage}[b]{0.72\linewidth}
		\subfloat[HR \protect\\ PSNR / SSIM]{
			\centering
			\includegraphics[width=0.2\linewidth]{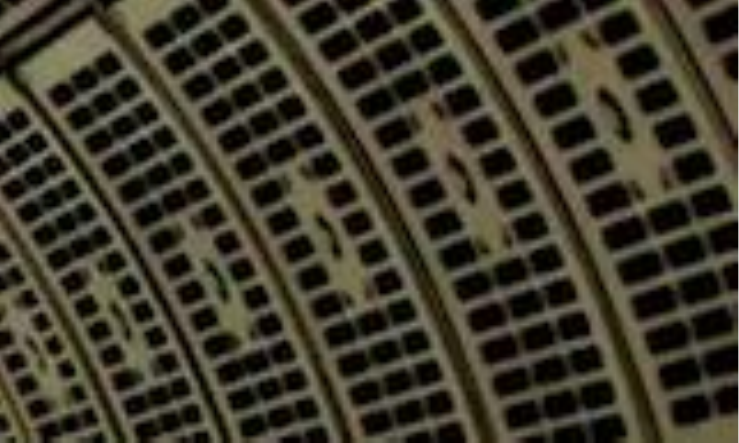}
		}
		\subfloat[Bicubic \protect\\ 25.70 / 0.6787]{
			\centering
			\includegraphics[width=0.2\linewidth]{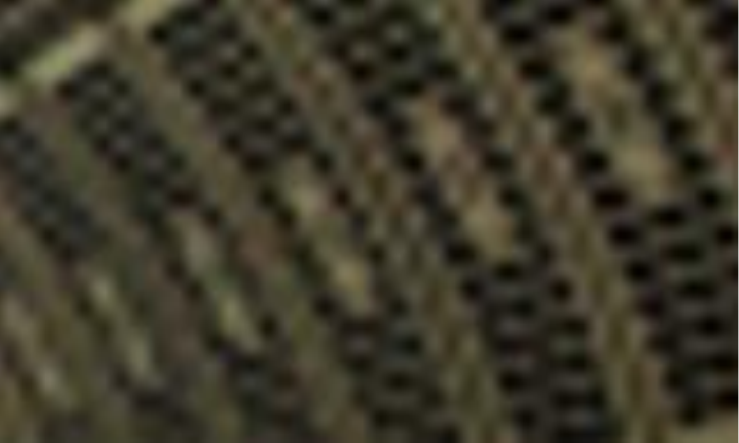}
		}
		\subfloat[MemNet~\cite{TaiY2017ICCV} \protect\\ 26.96 / 0.7628]{
			\centering
			\includegraphics[width=0.2\linewidth]{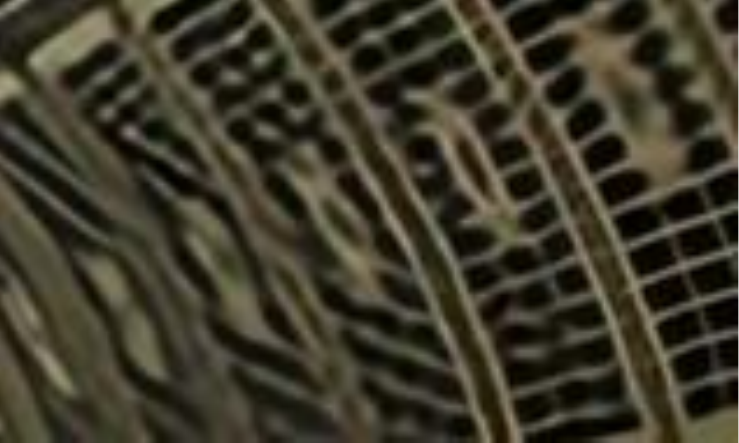}
		}
		\subfloat[EDSR~\cite{LimB2017CVPRW} \protect\\ 27.89 / 0.7938]{
			\centering
			\includegraphics[width=0.2\linewidth]{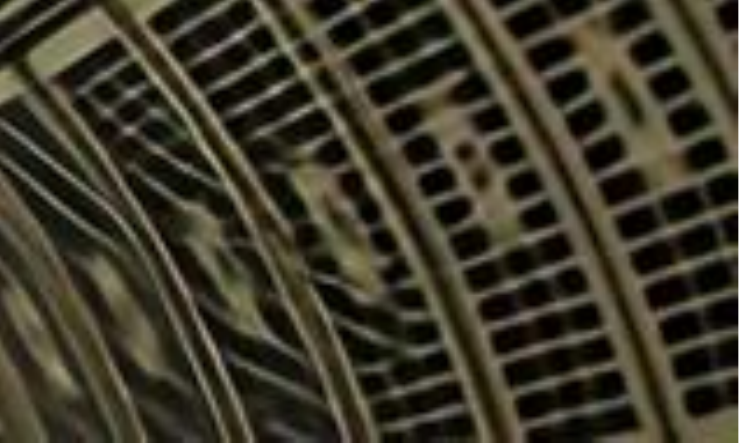}
		}
		\subfloat[RDN~\cite{ZhangY2018CVPR} \protect\\ 27.96 / 0.7938]{
			\centering
			\includegraphics[width=0.2\linewidth]{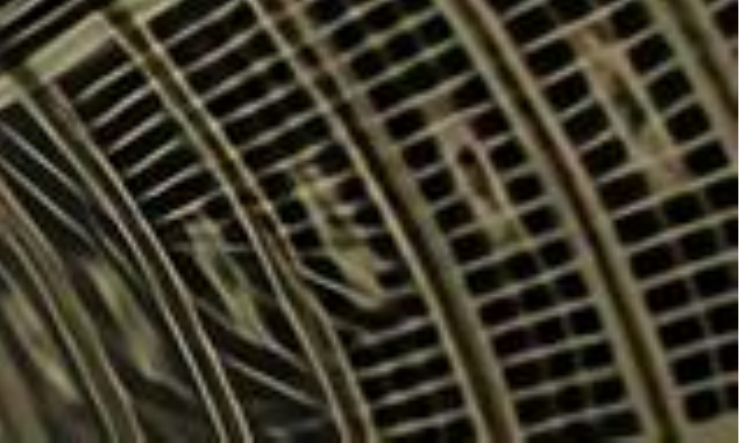}
		}\\
		\subfloat[MSRN~\cite{LiJ2018ECCV} \protect\\ 27.42 / 0.7796]{
			\centering
			\includegraphics[width=0.2\linewidth]{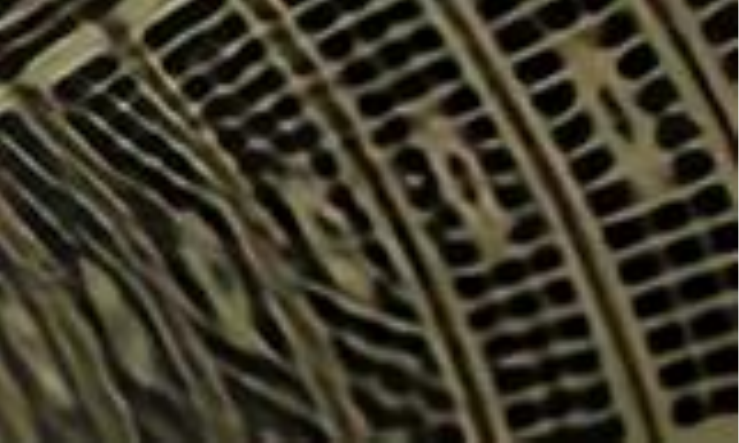}
		}
		\subfloat[RCAN\cite{ZhangY2018ECCV} \protect\\ 28.40 / 0.8030]{
			\centering
			\includegraphics[width=0.2\linewidth]{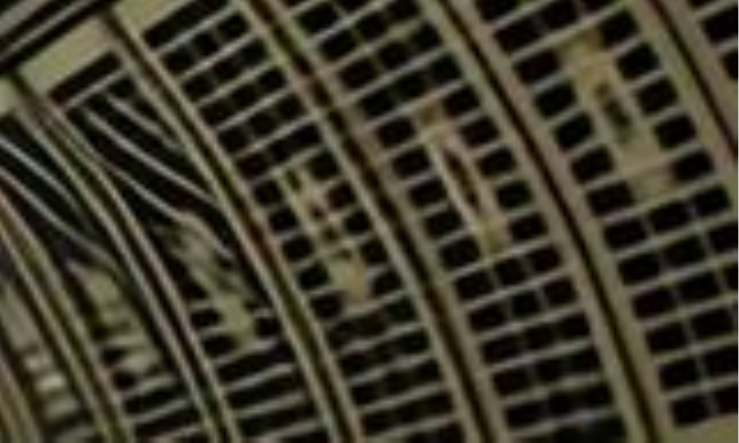}
		}
		\subfloat[CSFM~\cite{HuY2019TCSVT} \protect\\ 28.46 / 0.8017]{
			\centering
			\includegraphics[width=0.2\linewidth]{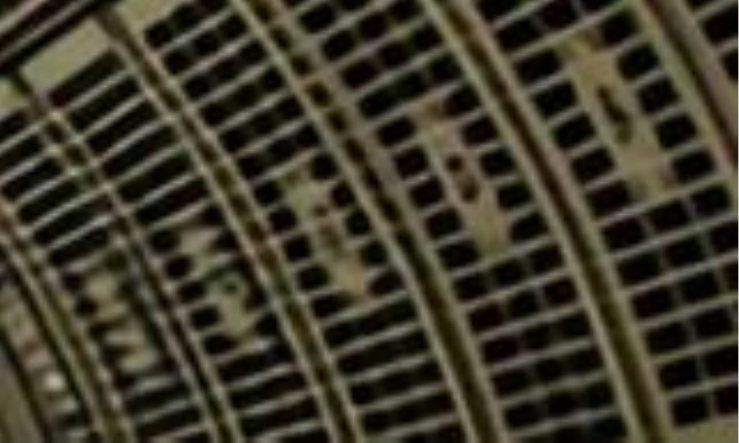}
		}
		\subfloat[OISR-RK3~\cite{HeX2019CVPR} \protect\\ 28.09 / 0.7962]{
			\centering
			\includegraphics[width=0.2\linewidth]{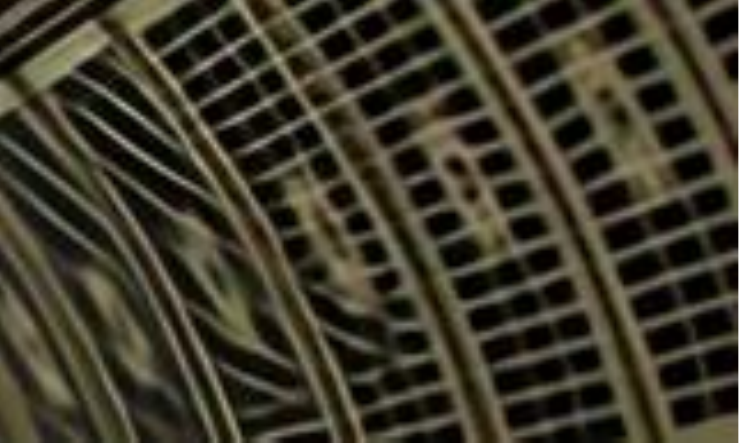}
		}
		\subfloat[$\textit{DeFiAN}_L$(Ours) \protect\\ {\bf28.69} / {\bf0.8055}]{
			\centering
			\includegraphics[width=0.2\linewidth]{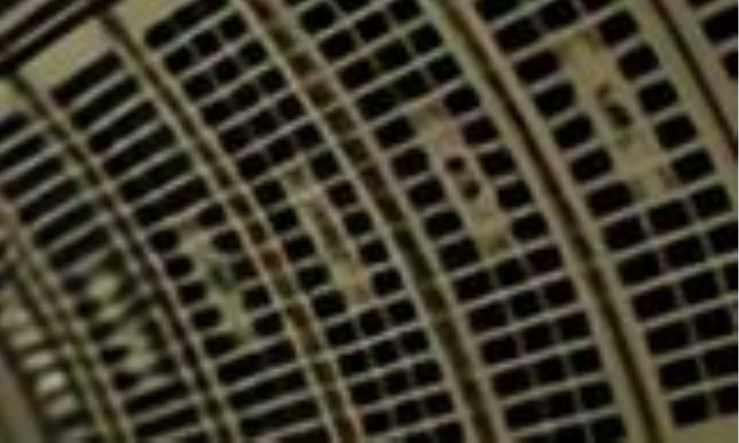}
		}
	\end{minipage}
	\caption{Subjective quality assessment for $\times4$ upscaling on "{\em img078}" from Urban100 dataset. Structural textures (\eg, windows) in our $\textit{DeFiAN}_L$ are more distinct than other state-of-the-art high-fidelity SISR methods.}
	\label{fig:Subjective comparisons:L1}
\end{figure*}
\begin{figure*}
    \captionsetup[subfloat]{justification=centering}
	\subfloat[Ground truth HR \protect\\ "{\em MomoyamaHaikagura}"]{
		\centering
		\includegraphics[width=0.208\linewidth]{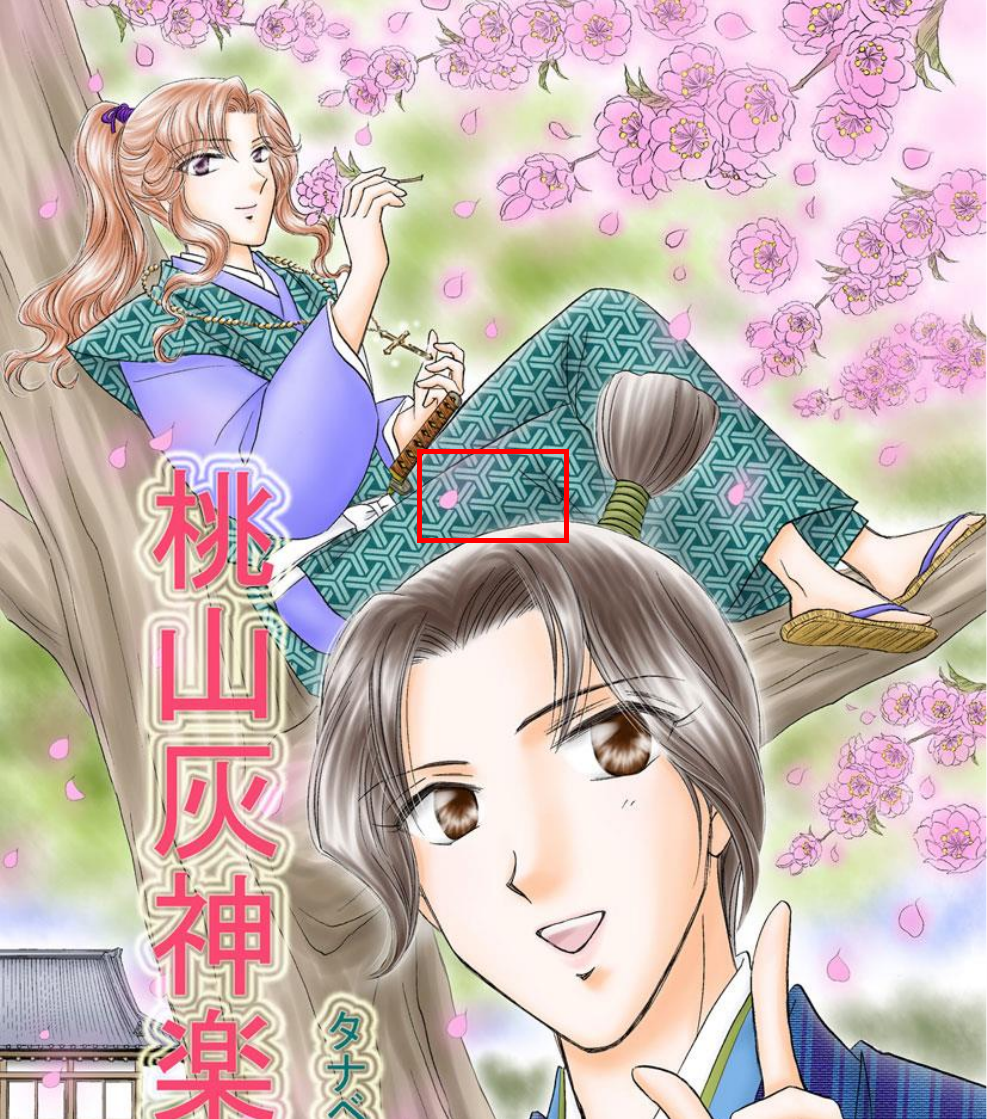}
	}
	\begin{minipage}[b]{0.72\linewidth}
		\subfloat[HR \protect\\ PSNR / SSIM]{
			\centering
			\includegraphics[width=0.2\linewidth]{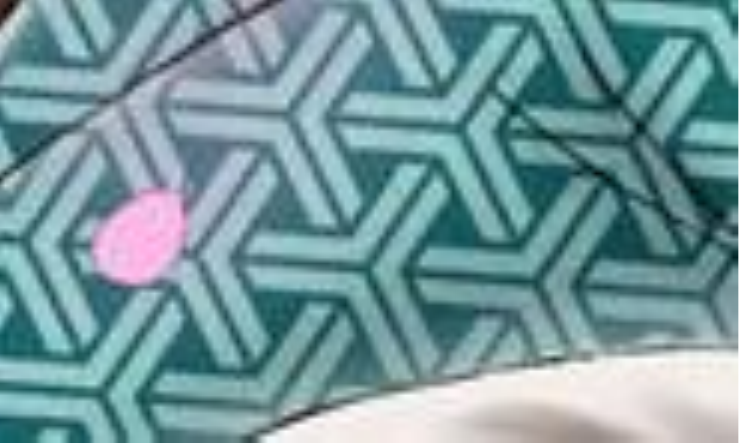}
		}
		\subfloat[Bicubic \protect\\ 21.62 / 0.6180]{
			\centering
			\includegraphics[width=0.2\linewidth]{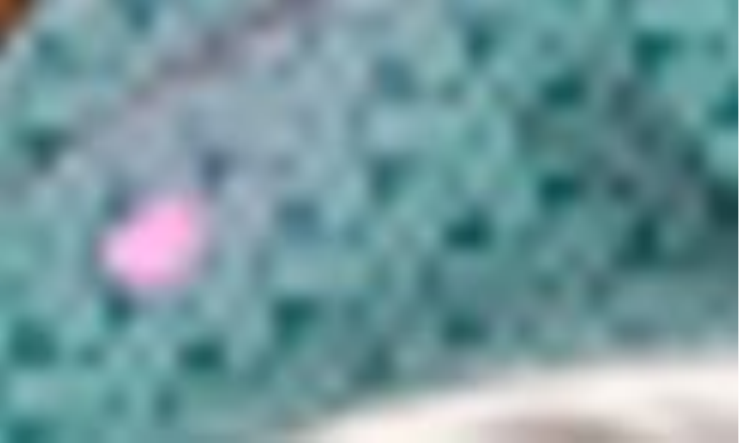}
		}
		\subfloat[MemNet~\cite{TaiY2017ICCV} \protect\\ 23.30 / 0.7380]{
			\centering
			\includegraphics[width=0.2\linewidth]{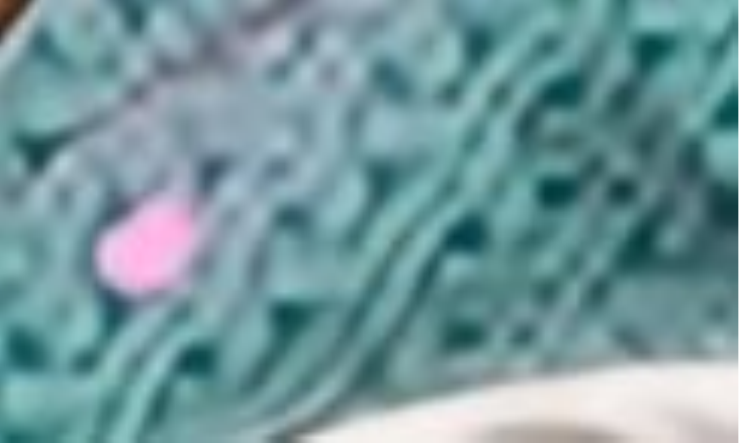}
		}
		\subfloat[EDSR~\cite{LimB2017CVPRW} \protect\\ 24.00 / 0.7802]{
			\centering
			\includegraphics[width=0.2\linewidth]{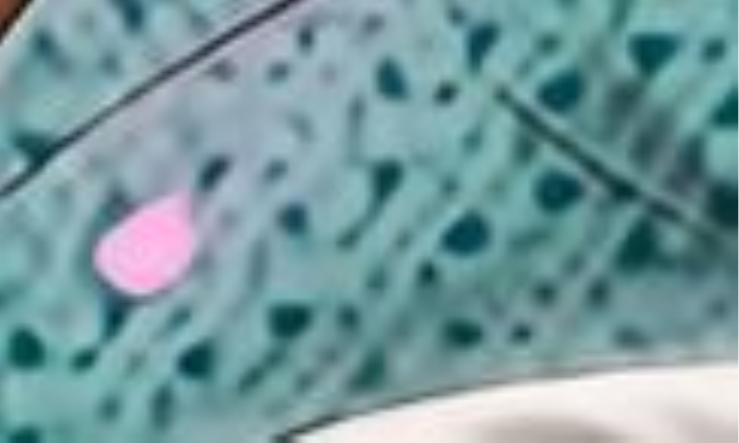}
		}
		\subfloat[RDN~\cite{ZhangY2018CVPR} \protect\\ 23.98 / 0.7812]{
			\centering
			\includegraphics[width=0.2\linewidth]{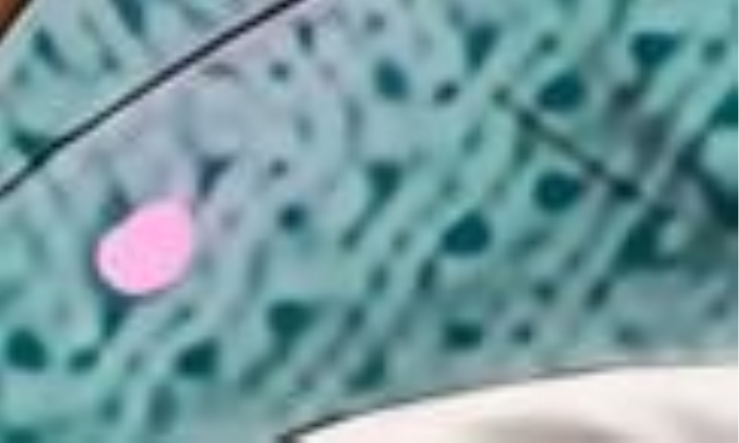}
		}\\
		\subfloat[MSRN~\cite{LiJ2018ECCV} \protect\\ 23.69 / 0.7627]{
			\centering
			\includegraphics[width=0.2\linewidth]{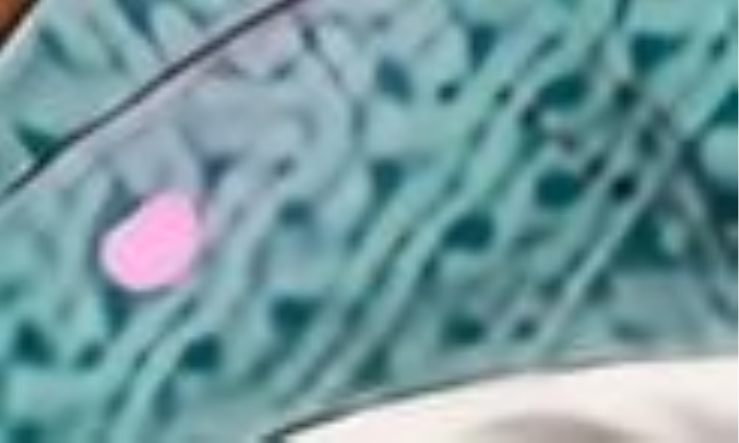}
		}
		\subfloat[RCAN\cite{ZhangY2018ECCV} \protect\\ 23.84 / 0.7833]{
			\centering
			\includegraphics[width=0.2\linewidth]{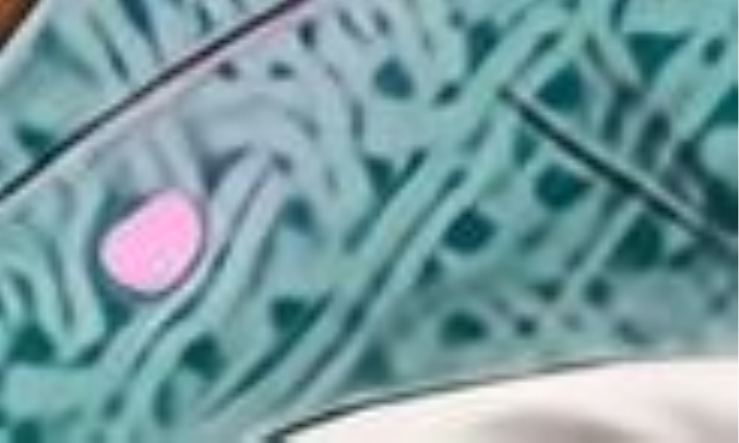}
		}
		\subfloat[CSFM~\cite{HuY2019TCSVT} \protect\\ 23.87 / 0.7856]{
			\centering
			\includegraphics[width=0.2\linewidth]{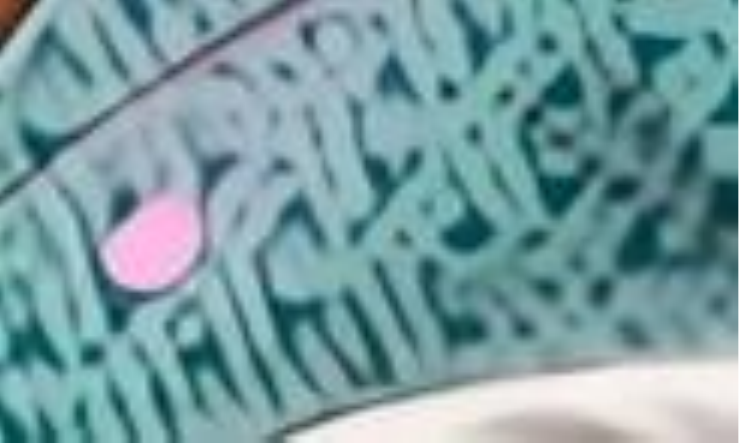}
		}
		\subfloat[OISR-RK3~\cite{HeX2019CVPR} \protect\\ 24.15 / 0.7869]{
			\centering
			\includegraphics[width=0.2\linewidth]{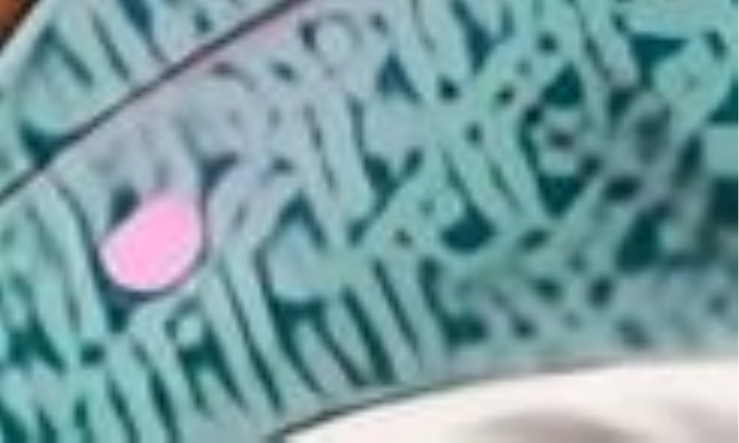}
		}
		\subfloat[$\textit{DeFiAN}_L$(Ours) \protect\\ {\bf24.17} / {\bf0.7925}]{
			\centering
			\includegraphics[width=0.2\linewidth]{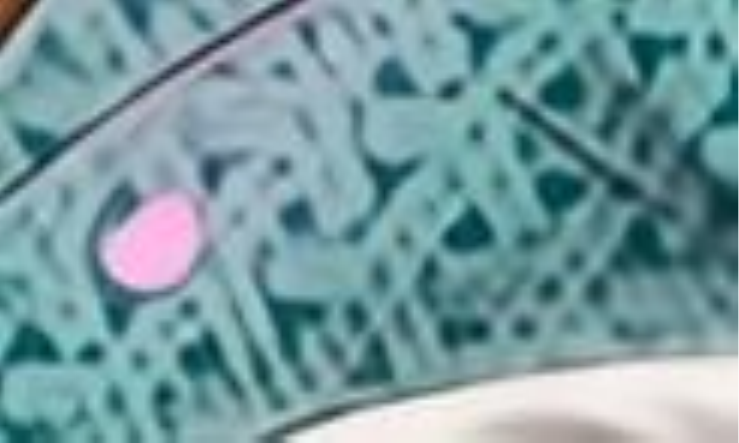}
		}
	\end{minipage}
	\caption{Subjective quality assessment for $\times4$ upscaling on "{\em MomoyamaHaikagura}" from Manga109 dataset. Our $\textit{DeFiAN}_L$ reconstruct the repetitive patterns better than other state-of-the-art high-fidelity SISR methods.}
	\label{fig:Subjective comparisons:L2}
\end{figure*}
\subsection{Implementation Details}
\subsubsection{Hyperparameters}~\label{sec:Hyperparameters}
As shown in Fig.~\ref{fig:DeFiAN}, a full {\em DeFiAN} model depends on several hyperparameters, including the number of {\em DeFiAM} modules ($N$), the number of RCAB blocks in each $\textit{FEM}$ ($M$), the number of channels in each layer ($C$), and the specific settings of attention module. For the purposes on high-fidelity and lightweight applications, we propose two {\em DeFiAN} models:

i) High-fidelity ${\textit{DeFiAN}}_{L}$ has a chain of $N=10$ {\em DeFiAM} modules, and in each module, $M=20$ RCABs are stacked to reconstruct the primitive residues with $C=64$ channels.

ii) Lightweight ${\textit{DeFiAN}}_{S}$ has a chain of $N=5$ {\em DeFiAM} modules, and in each module, $M=10$ RCABs are stacked to reconstruct the primitive residues with $C=32$ channels.

Moreover, as described in Section~\ref{sec:DeFiAN}, for both ${\textit{DeFiAN}}_{L}$ and ${\textit{DeFiAN}}_{S}$, aiming at inferring the detail-fidelity attention representations for feature enhancement, the primitive residues are delivered into the subsequent modules, including the interpretable {\em MSHF} which aggregates three Hessian eigenvalues [$\boldsymbol{\bar{\lambda}}_{3}, \boldsymbol{\bar{\lambda}}_{5}, \boldsymbol{\bar{\lambda}}_{7}$] to infer the fine (edges and textures) and coarse (structures) details of feature, the {\em DiEnDec} with 3 dilated convolution layers for feature erosion and 3 dilated deconvolution layers for feature dilation as Fig.~\ref{fig:DiEnDec}(a) shows, and the {\em DAC} with 2 full connection layers for parameterizing the distribution of referenced features.

\subsubsection{Optimization}
All of our models are optimized via minimizing the mean absolute error (MAE) between the super-resolved image $\textit{DeFiAN}_*(\boldsymbol{x})$ and the corresponding groundtruth image $\boldsymbol{y}$. Therefore, given a training dataset $\{\boldsymbol{x}^k, \boldsymbol{y}^k\}^K_{k=1}$, where $K$ is the batch size and $\{\boldsymbol{x}^k, \boldsymbol{y}^k\}$ are the $k$-th LR and HR patch pairs. Then, the objective function is formulated as
\begin{equation}
\boldsymbol{\mathcal{L}}(\theta)=\sum_{k=1}^{K}\left \| \boldsymbol{y}^k - \textit{DeFiAN}(\boldsymbol{x}^k)  \right \|_1
\label{eq:objective function}
\end{equation}
where $\theta$ denotes the parameters of $\textit{DeFiAN}$ models, which are optimized using the Adam optimizer~\cite{KingmaD2014ICLR} with mini-batch of size $K=16$. The learning rate is initialized to ${{10}^{-4}}$ and halved for every $2\times10^5$ mini-batch updates. Each of the final models will get convergence after $6\times10^5$ mini-batch updates on PyTorch framework and a 12GB NVIDIA Titan X Pascal GPU.
\begin{table*}
	\centering
    \captionsetup{justification=centering}
	\caption{\textsc{Quantitative comparisons of the proposed $\textit{DeFiAN}_S$ with the state-of-the-art \\lightweight methods on benchmark datasets for $\times2$, $\times3$ and $\times4$ upscaling.}}
	\begin{tabular}{{c c c c c c c c c c c c c c}}
	\hline
    \hline
    \multirow{2}{*}{Scale}&\multirow{2}{*}{Method}&\#Params&\#FLOPs
    &\multicolumn{2}{c}{Set5}&\multicolumn{2}{c}{Set14}&\multicolumn{2}{c}{BSD100}
    &\multicolumn{2}{c}{Urban100}&\multicolumn{2}{c}{Manga109}\\
    &&(K)&(G)&PSNR&SSIM&PSNR&SSIM&PSNR&SSIM&PSNR&SSIM&PSNR&SSIM\\
	\hline
	\multirow{9}{*}{$\times2$}
    &Bicubic&-&-
    &33.64&0.9292&30.22&0.8683&29.55&0.8425&26.87&0.8397&30.80&0.9339\\
	&SRCNN~\cite{DongC2016TPAMI}&8.0&1.4
    &36.66&0.9542&32.47&0.9069&31.37&0.8879&29.52&0.8947&35.73&0.9675\\
	&FSRCNN\cite{DongC2016ECCV}&12.5&0.5
    &36.98&0.9556&32.62&0.9087&31.50&0.8904&29.85&0.9009&36.56&0.9703\\
    &VDSR~\cite{KimJ2016CVPR_VDSR}&664.7&115.1
    &37.53&0.9587&33.05&0.9127&31.90&0.8960&30.77&0.9141&37.37&0.9737\\
	&LapSRN~\cite{LaiW2017CVPR}&435.3&18.9
    &37.52&0.9591&32.99&0.9124&31.80&0.8949&30.41&0.9101&37.27&0.9740\\
	&DRRN~\cite{TaiY2017CVPR}&297.2&1275.5
    &37.74&0.9591&33.25&0.9137&32.05&0.8973&31.23&0.9188&37.88&0.9750\\
	&IDN~\cite{HuiZ2018CVPR}&715.3&31.0
    &\underline{37.83}&\underline{0.9600}&33.30&0.9148&32.08&\underline{0.8985}&31.27&0.9196&38.02&0.9749\\
	&CARN~\cite{AhnN2018ECCV}&1592.4&41.9
    &37.80&0.9589&\underline{33.44}&\underline{0.9161}&\underline{32.10}&0.8978&\underline{31.93}
    &\underline{0.9256}&\underline{38.31}&\underline{0.9754}\\
	&$\textit{DeFiAN}_S$(Ours)&1027.6&44.2
    &{\bf38.03}&{\bf0.9605}&{\bf33.63}&{\bf0.9181}&{\bf32.20}&{\bf0.8999}&{\bf32.20}&{\bf0.9286}&{\bf38.91}&{\bf0.9775}\\
	\hline
	\multirow{9}{*}{$\times3$}
    &Bicubic&-&-
    &30.40&0.8686&27.54&0.7741&27.21&0.7389&24.46&0.7349&26.95&0.8565\\
	&SRCNN~\cite{DongC2016TPAMI}&8.0&1.4
    &32.75&0.9090&29.29&0.8215&28.41&0.7863&26.24&0.7991&30.56&0.9125\\
	&FSRCNN\cite{DongC2016ECCV}&12.5&0.2
    &33.16&0.9140&29.42&0.8242&28.52&0.7893&26.41&0.8064&31.12&0.9196\\
	&VDSR~\cite{KimJ2016CVPR_VDSR}&664.7&115.1
    &33.66&0.9213&29.78&0.8318&28.83&0.7976&27.14&0.8279&32.13&0.9348\\
	&LapSRN~\cite{LaiW2017CVPR}&435.3&8.5
    &33.82&0.9227&29.79&0.8320&28.82&0.7973&27.07&0.8271&32.21&0.9344\\
	&DRRN~\cite{TaiY2017CVPR}&297.2&1275.5
    &34.02&0.9244&29.98&0.8350&28.95&0.8004&27.54&0.8378&32.72&0.9380\\
	&IDN~\cite{HuiZ2018CVPR}&715.3&13.8
    &34.12&0.9253&29.99&0.8356&28.95&0.8013&27.42&0.8360&32.71&0.9379\\
	&CARN~\cite{AhnN2018ECCV}&1592.4&22.3
    &\underline{34.28}&\underline{0.9254}&\underline{30.19}&\underline{0.8397}&\underline{29.06}
    &\underline{0.8034}&\underline{28.06}&\underline{0.8493}&\underline{33.46}&\underline{0.9429}\\
	&$\textit{DeFiAN}_S$(Ours)&1073.7&20.6
    &\bf{34.42}&\bf{0.9273}&\bf{30.34}&\bf{0.8410}&\bf{29.12}&\bf{0.8053}&\bf{28.20}&\bf{0.8528}&\bf{33.72}&\bf{0.9447}\\
	\hline
	\multirow{9}{*}{$\times4$}
    &Bicubic&-&-
    &28.42&0.8101&25.99&0.7023&25.96&0.6672&23.14&0.6573&24.89&0.7866\\
	&SRCNN~\cite{DongC2016TPAMI}&8.0&1.4
    &30.49&0.8629&27.51&0.7519&26.91&0.7104&24.53&0.7230&27.66&0.8566\\
	&FSRCNN\cite{DongC2016ECCV}&12.5&0.1
    &30.70&0.8657&27.59&0.7535&26.96&0.7128&24.60&0.7258&27.85&0.8557\\
	&VDSR~\cite{KimJ2016CVPR_VDSR}&664.7&115.1
    &31.35&0.8838&28.02&0.7678&27.29&0.7252&25.18&0.7525&28.87&0.8865\\
	&LapSRN~\cite{LaiW2017CVPR}&435.3&4.8
    &31.52&0.8854&28.09&0.7687&27.31&0.7255&25.21&0.7545&29.08&0.8883\\
	&DRRN~\cite{TaiY2017CVPR}&297.2&1275.5
    &31.67&0.8888&28.22&0.7721&27.38&0.7284&25.45&0.7639&29.44&0.8943\\
	&IDN~\cite{HuiZ2018CVPR}&715.3&7.7
    &31.81&0.8903&28.25&0.7731&27.41&0.7295&25.41&0.7630&29.42&0.8939\\
	&CARN~\cite{AhnN2018ECCV}&1592.4&17.1
    &\underline{32.12}&\underline{0.8936}&\underline{28.50}&\underline{0.7791}&\bf{27.58}&\underline{0.7349}
    &\underline{26.06}&\underline{0.7837}&\underline{30.45}&\underline{0.9073}\\ 
	&$\textit{DeFiAN}_S$(Ours)&1064.5&12.8
    &\bf{32.16}&\bf{0.8942}&\bf{28.63}&\bf{0.7810}&{\bf27.58}&\bf{0.7363}&\bf{26.10}&\bf{0.7862}&{\bf30.59}&{\bf0.9084}\\
	\hline
    \hline
	\end{tabular}
	\label{tab:full-reference_comparison_lightweight}
\end{table*}
\begin{figure*}
    \captionsetup[subfloat]{justification=centering}
	\subfloat[Ground truth HR \protect\\ "{\em img092}"]{
		\centering
		\includegraphics[width=0.208\linewidth]{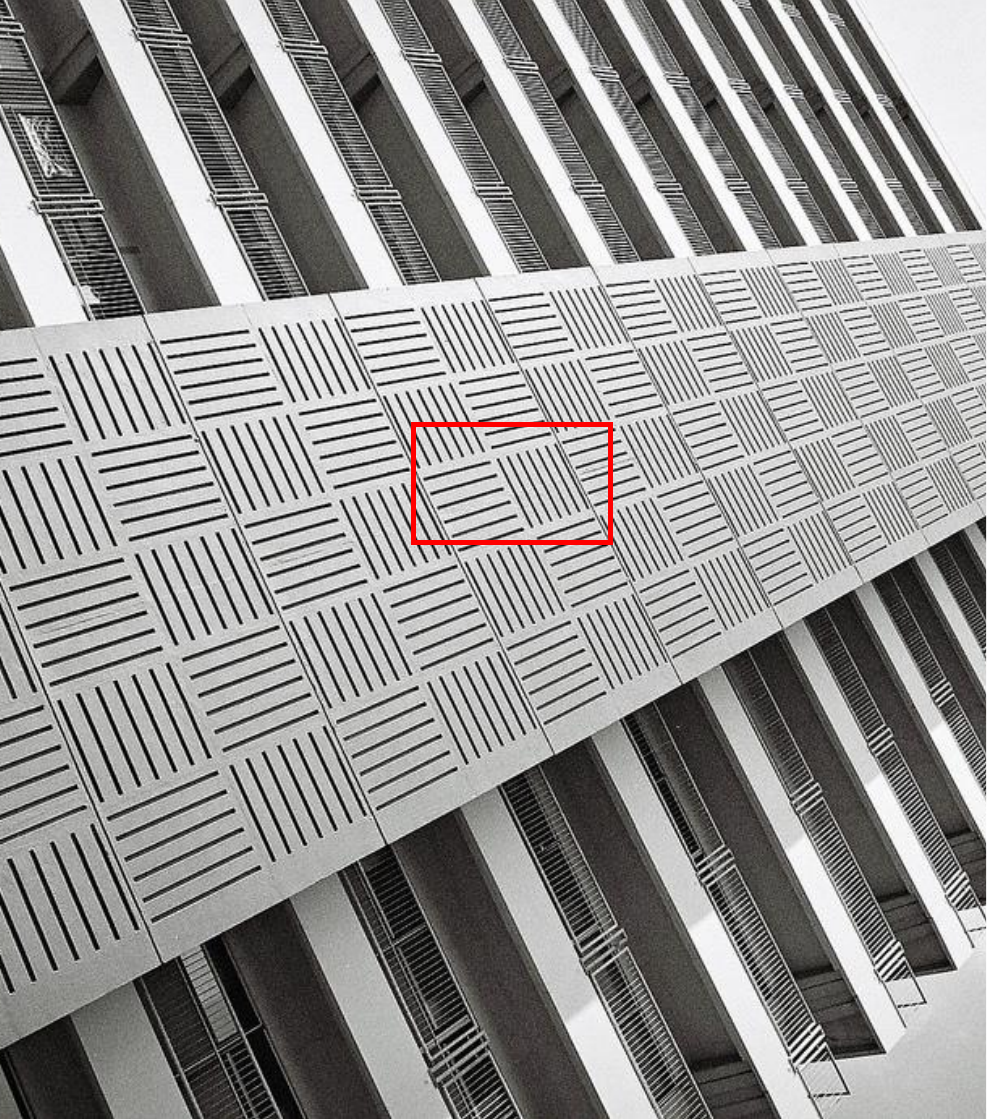}
	}
	\begin{minipage}[b]{0.72\linewidth}
		\subfloat[HR \protect\\ PSNR / SSIM]{
			\centering
			\includegraphics[width=0.2\linewidth]{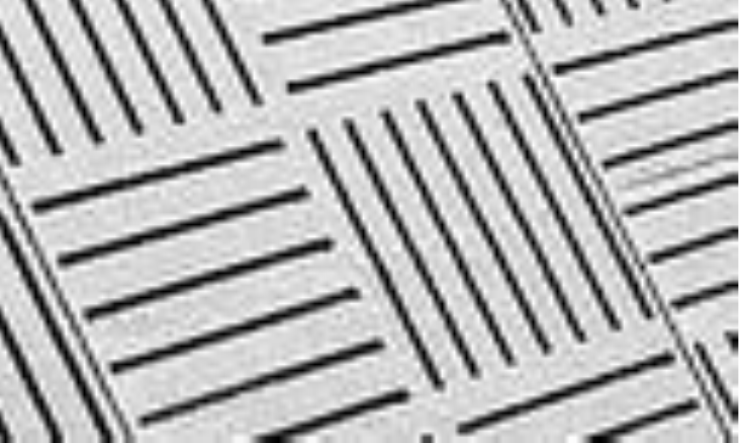}
		}
		\subfloat[Bicubic \protect\\ 16.58 / 0.4371]{
			\centering
			\includegraphics[width=0.2\linewidth]{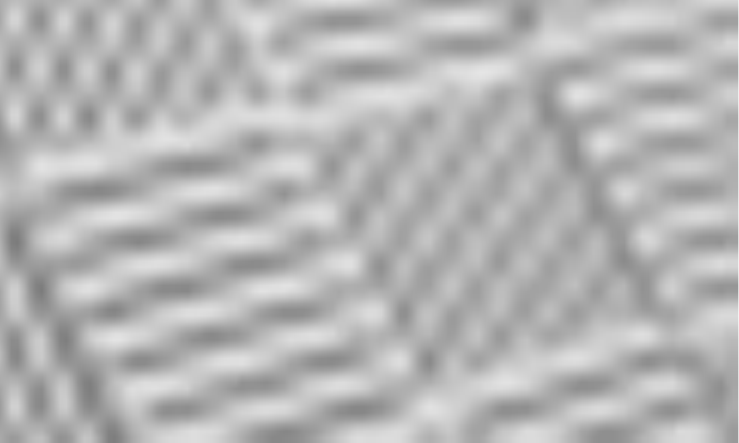}
		}
		\subfloat[SRCNN~\cite{DongC2016TPAMI} \protect\\ 17.56 / 0.5414]{
			\centering
			\includegraphics[width=0.2\linewidth]{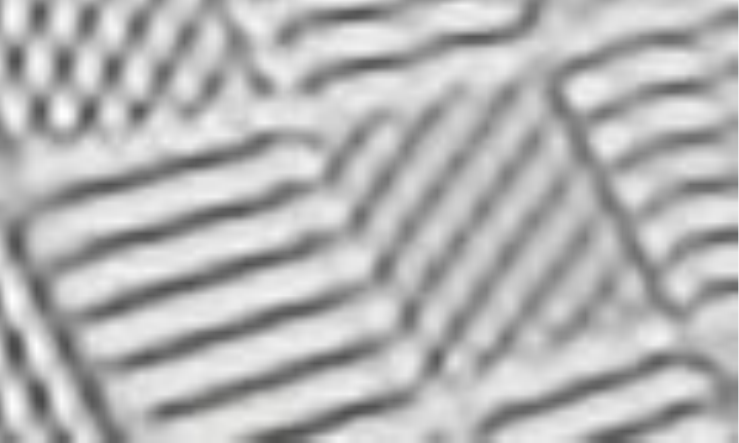}
		}
		\subfloat[FSRCNN~\cite{DongC2016ECCV} \protect\\ 17.72 / 0.5671]{
			\centering
			\includegraphics[width=0.2\linewidth]{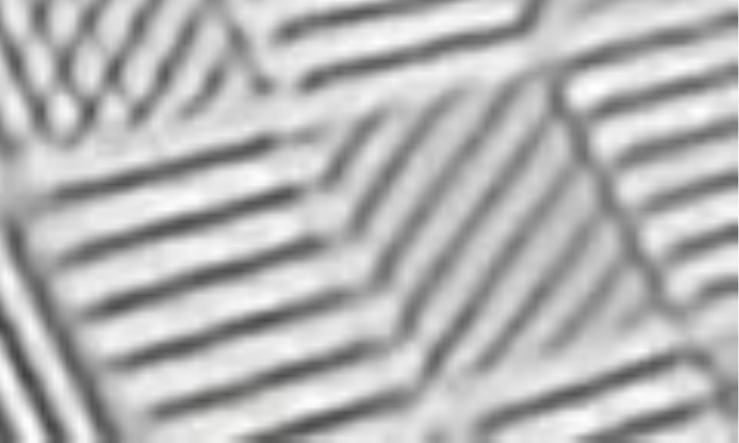}
		}
		\subfloat[VDSR~\cite{KimJ2016CVPR_VDSR} \protect\\ 18.14 / 0.6011]{
			\centering
			\includegraphics[width=0.2\linewidth]{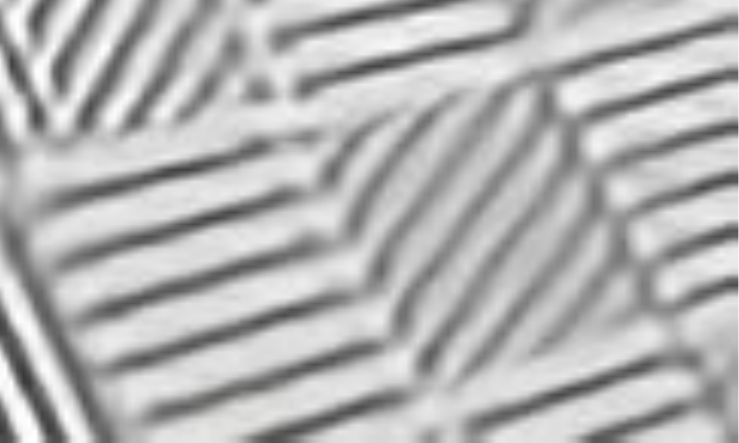}
		}\\
		\subfloat[LapSRN~\cite{LaiW2017CVPR} \protect\\ 18.20 / 0.6073]{
			\centering
			\includegraphics[width=0.2\linewidth]{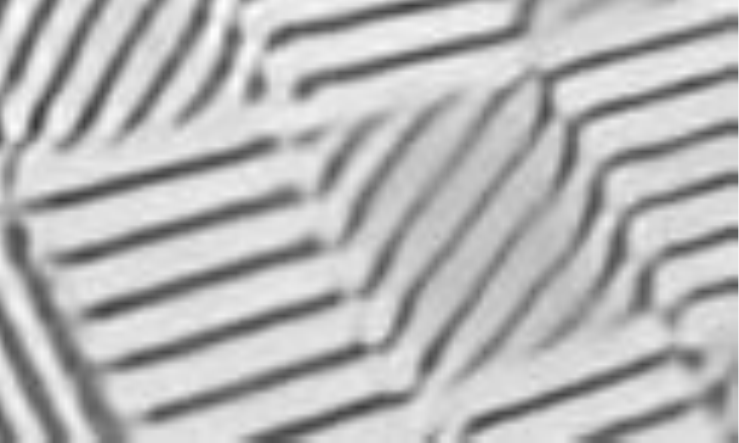}
		}
		\subfloat[DRRN\cite{TaiY2017CVPR} \protect\\ 18.58 / 0.6373]{
			\centering
			\includegraphics[width=0.2\linewidth]{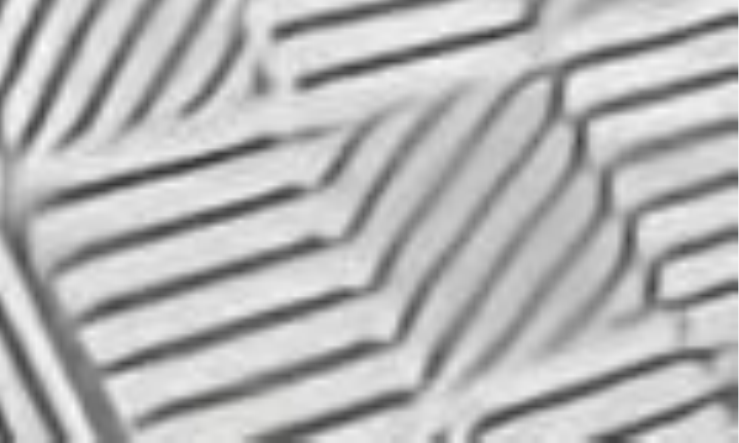}
		}
		\subfloat[IDN~\cite{HuiZ2018CVPR} \protect\\ 18.27 / 0.6176]{
			\centering
			\includegraphics[width=0.2\linewidth]{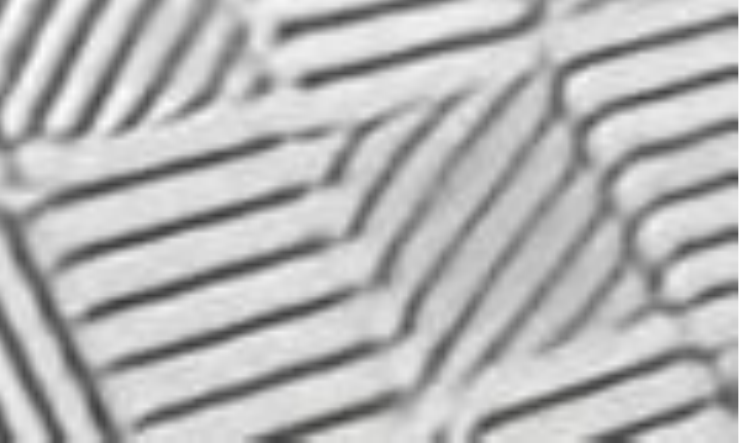}
		}
		\subfloat[CARN~\cite{AhnN2018ECCV} \protect\\ 18.91 / 0.6585]{
			\centering
			\includegraphics[width=0.2\linewidth]{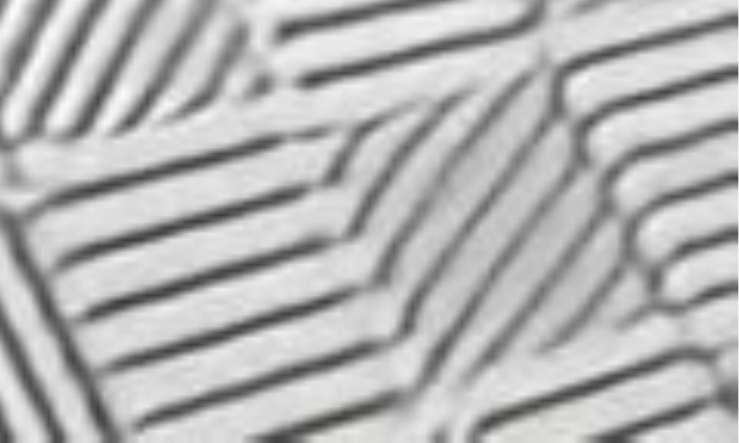}
		}
		\subfloat[$\textit{DeFiAN}_S$(Ours) \protect\\ {\bf19.33} / {\bf0.6706}]{
			\centering
			\includegraphics[width=0.2\linewidth]{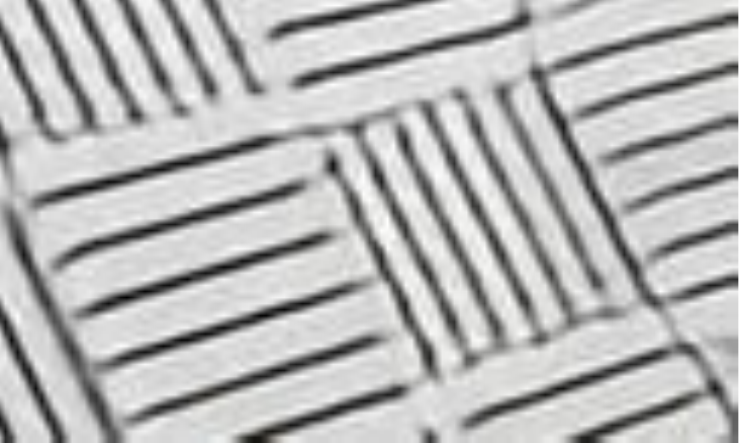}
		}
	\end{minipage}
	\caption{Subjective quality assessment for $\times4$ upscaling on "{\em img092}" from Urban100 dataset. Only our $\textit{DeFiAN}_S$ can reconstruct the mixed edges correctly which are failed in other state-of-the-art lightweight SISR methods.}
	\label{fig:Subjective comparisons:S1}
\end{figure*}

\begin{figure*}
    \captionsetup[subfloat]{justification=centering}
	\subfloat[Ground truth HR \protect\\ "{\em GOOD\_KISS\_Ver2}"]{
		\centering
		\includegraphics[width=0.208\linewidth]{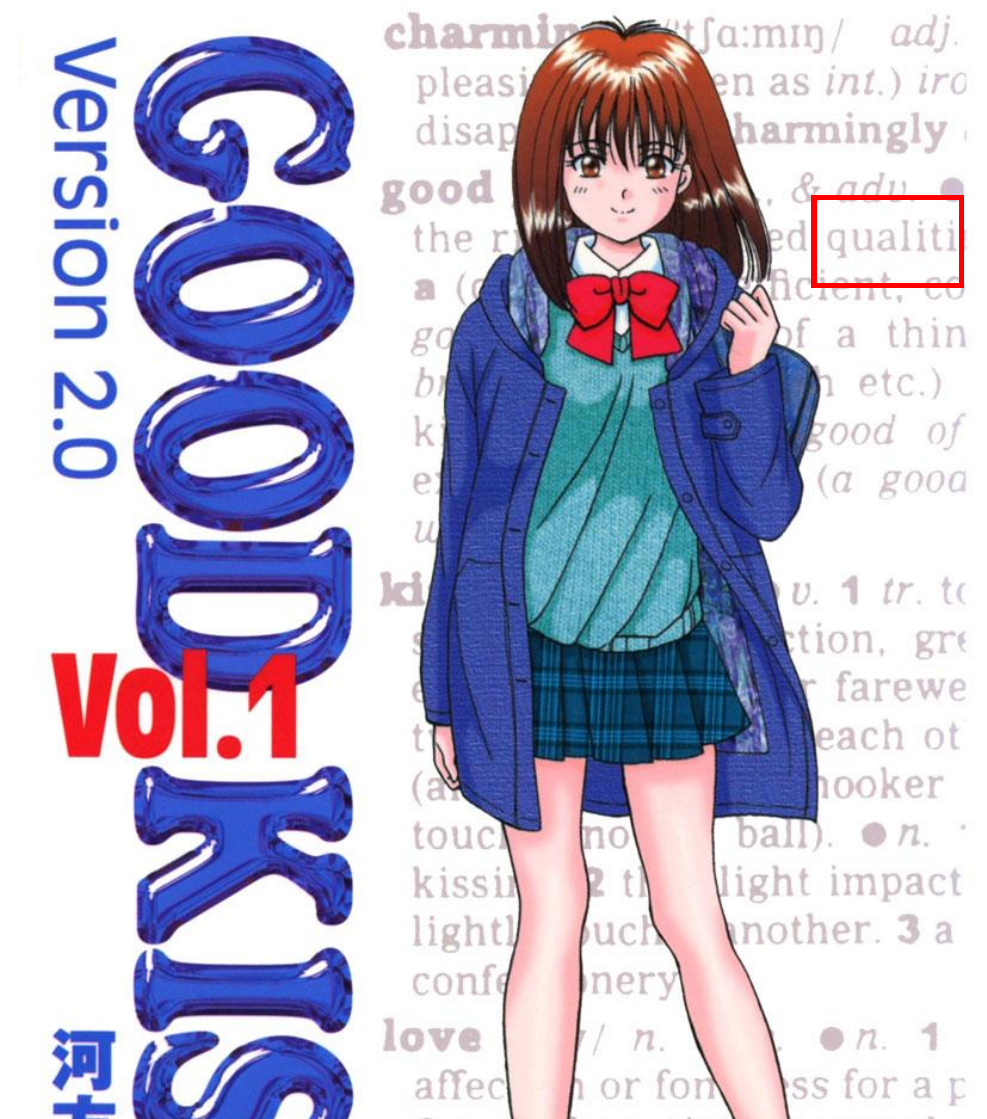}
	}
	\begin{minipage}[b]{0.72\linewidth}
		\subfloat[HR \protect\\ PSNR / SSIM]{
			\centering
			\includegraphics[width=0.2\linewidth]{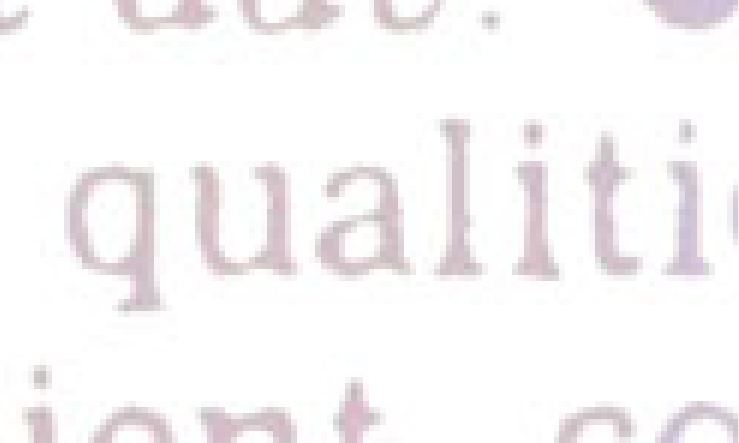}
		}
		\subfloat[Bicubic \protect\\ 25.52 / 0.7984]{
			\centering
			\includegraphics[width=0.2\linewidth]{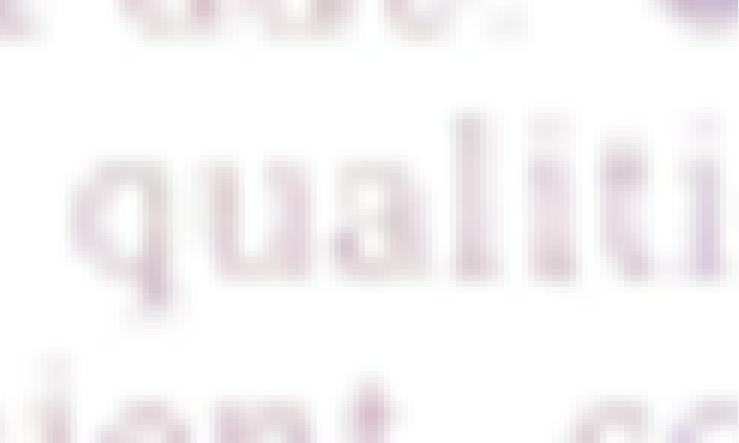}
		}
		\subfloat[SRCNN~\cite{DongC2016TPAMI} \protect\\ 28.98 / 0.8751]{
			\centering
			\includegraphics[width=0.2\linewidth]{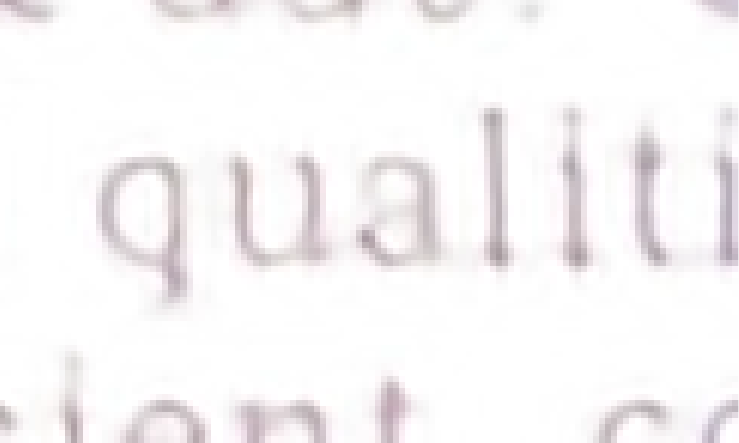}
		}
		\subfloat[FSRCNN~\cite{DongC2016ECCV} \protect\\ 30.31 / 0.8959]{
			\centering
			\includegraphics[width=0.2\linewidth]{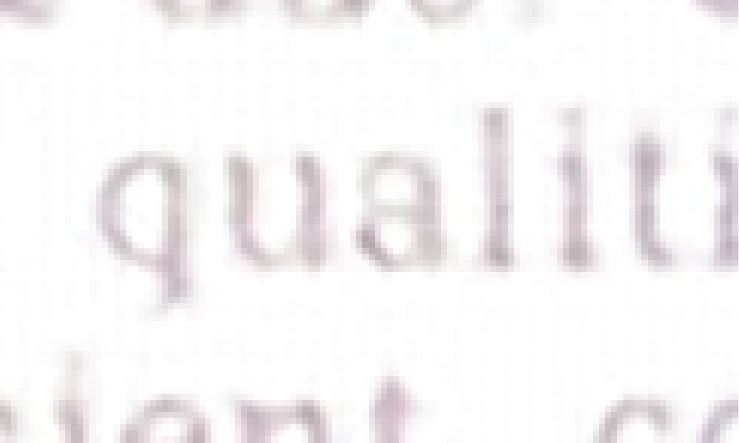}
		}
		\subfloat[VDSR~\cite{KimJ2016CVPR_VDSR} \protect\\ 30.55 / 0.9030]{
			\centering
			\includegraphics[width=0.2\linewidth]{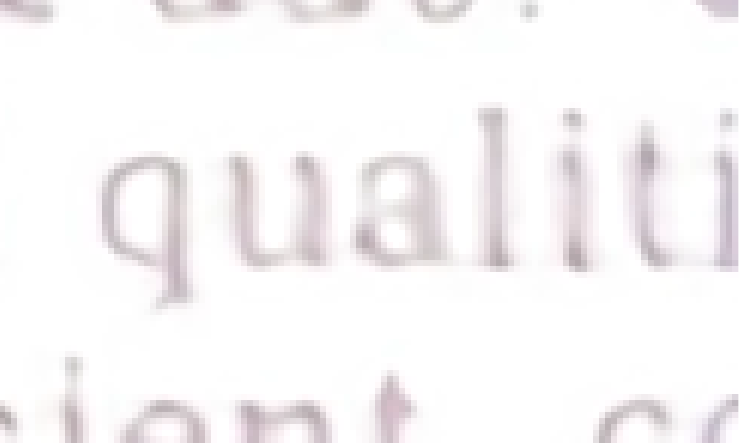}
		}\\
		\subfloat[LapSRN~\cite{LaiW2017CVPR} \protect\\ 30.59 / 0.9047]{
			\centering
			\includegraphics[width=0.2\linewidth]{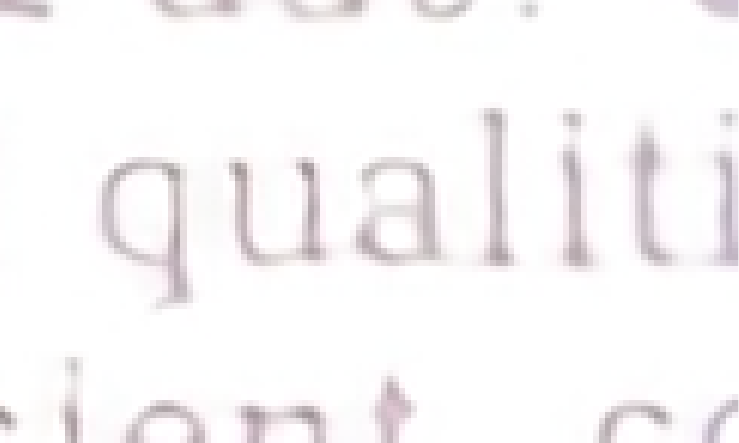}
		}
		\subfloat[DRRN\cite{TaiY2017CVPR} \protect\\ 31.10 / 0.9108]{
			\centering
			\includegraphics[width=0.2\linewidth]{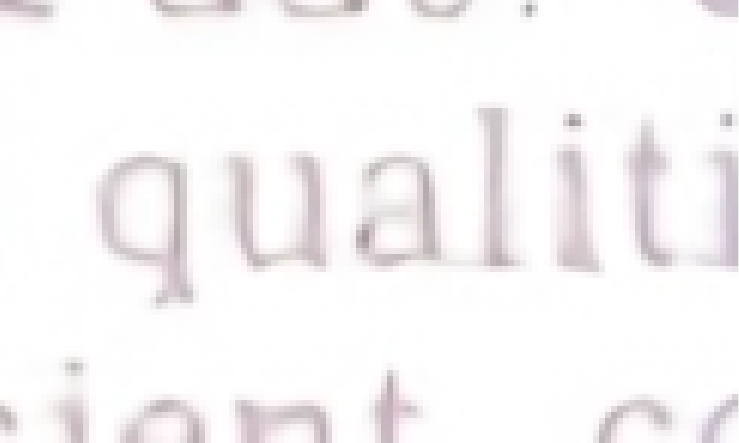}
		}
		\subfloat[IDN~\cite{HuiZ2018CVPR} \protect\\ 31.19 / 0.9110]{
			\centering
			\includegraphics[width=0.2\linewidth]{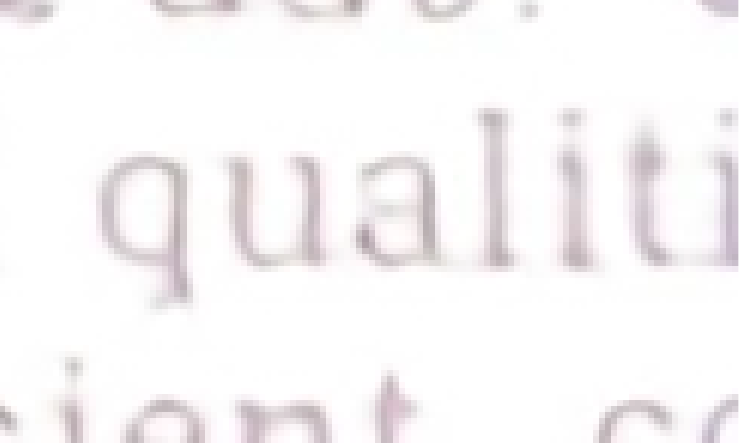}
		}
		\subfloat[CARN~\cite{AhnN2018ECCV} \protect\\ 31.75 / {\bf0.9206}]{
			\centering
			\includegraphics[width=0.2\linewidth]{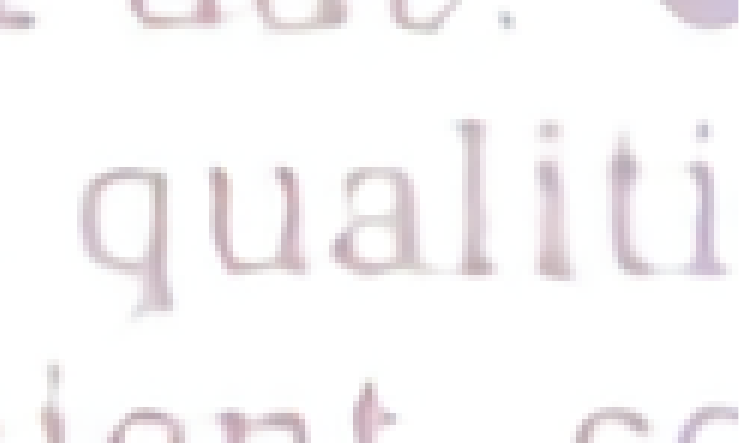}
		}
		\subfloat[$\textit{DeFiAN}_S$(Ours) \protect\\ {\bf31.89} / 0.9200]{
			\centering
			\includegraphics[width=0.2\linewidth]{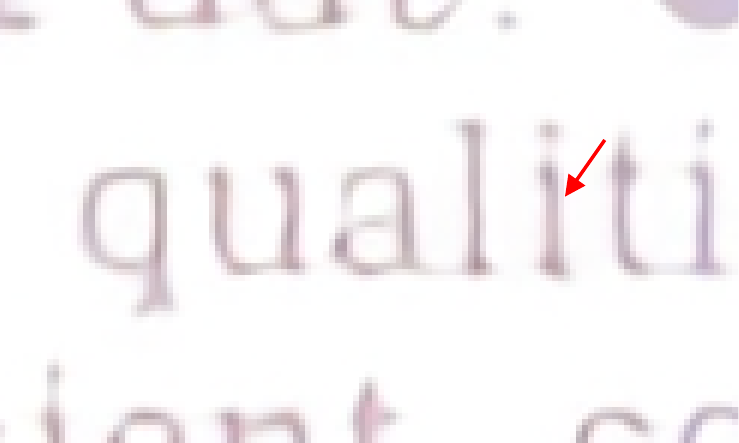}
		}
	\end{minipage}
	\caption{Subjective quality assessment for $\times4$ upscaling on "{\em GOOD\_KISS\_Ver2}" from Manga109 dataset. Our $\textit{DeFiAN}_S$ performs better than other state-of-the-art lightweight SISR methods on reconstructing the texts/characters.}
	\label{fig:Subjective comparisons:S2}
\end{figure*}
\subsection{Comparison with State-of-the-art Methods}
To illustrate the effectiveness of the proposed $\textit{DeFiAN}$ models, several state-of-the-art SISR methods\footnote[1]{To be fair, all the comparative experiments are conducted on the super-resolved byte-format images generated using the released codes: \href{http://mmlab.ie.cuhk.edu.hk/projects/SRCNN.html}{SRCNN}, \href{http://mmlab.ie.cuhk.edu.hk/projects/FSRCNN.html}{FSRCNN}, \href{https://cv.snu.ac.kr/research/VDSR/}{VDSR}, \href{http://vllab.ucmerced.edu/wlai24/LapSRN/}{LapSRN}, \href{https://github.com/tyshiwo/DRRN_CVPR17}{DRRN}, \href{https://github.com/tyshiwo/MemNet}{MemNet}, \href{https://github.com/Zheng222/IDN-Caffe}{IDN}, \href{https://github.com/LimBee/NTIRE2017}{EDSR}, \href{http://www.toyota-ti.ac.jp/Lab/Denshi/iim/members/muhammad.haris/projects/DBPN.html}{D-DBPN}, \href{https://github.com/yulunzhang/RDN}{RDN}, \href{https://github.com/yulunzhang/RCAN}{RCAN}, \href{https://github.com/MIVRC/MSRN-PyTorch}{MSRN}, \href{https://github.com/nmhkahn/CARN-pytorch}{CARN}, \href{https://github.com/HolmesShuan/OISR-PyTorch}{OISR}.} are compared in terms of quantitative objective evaluation, qualitatively visual performance and computational complexity. As mentioned in Section~\ref{sec:Hyperparameters}, for the purposes on high-fidelity and lightweight applications, two models $\textit{DeFiAN}_L$ and $\textit{DeFiAN}_S$ are proposed and compared with other corresponding state-of-the-art SISR methods.
\subsubsection{High-fidelity models}
Aiming at evaluating the high-fidelity super-resolved images from the low-resolution inputs, several state-of-the-art high-fidelity SR methods are selected to compare with our proposed $\textit{DeFiAN}_L$, including MemNet~\cite{TaiY2017ICCV}, EDSR~\cite{LimB2017CVPRW}, D-DBPN~\cite{Haris2018CVPR}, RDN~\cite{ZhangY2018CVPR}, MSRN~\cite{LiJ2018ECCV}, RCAN~\cite{ZhangY2018ECCV}, CSFM~\cite{HuY2019TCSVT} and OISR-RK3~\cite{HeX2019CVPR}. As posted in TABLE~\ref{tab:full-reference_comparison_highfidelity}, although compared with RCAN~\cite{ZhangY2018ECCV} and CSFM~\cite{HuY2019TCSVT}, both of which apply the attention mechanism into very deep networks, our $\textit{DeFiAN}_L$ achieves the highest performance with comparative spacial complexity and time complexity.

As described in Section~\ref{sec:DeFiAN}, since we stack a chain of residual channel attention blocks in the {\em FEM} modules, in a sense, RCAN~\cite{ZhangY2018ECCV} is considered as the baseline of our {\em DeFiAN} models. Particularly, by introducing the detail-fidelity attention mechanism, our {\em DeFiAN} is equipped with stronger ability of detail representation. As reported in TABLE~\ref{tab:full-reference_comparison_highfidelity}, especially on the challenging dataset Manga109, the proposed $\textit{DeFiAN}_L$ advances RCAN~\cite{ZhangY2018ECCV} with the improvement margins of 0.34 dB, 0.41 dB and 0.47 dB for $\times2$, $\times3$ and $\times4$ upscaling respectively. Besides, as shown in Fig.~\ref{fig:Subjective comparisons:L1}-\ref{fig:Subjective comparisons:L2}, our {\em DeFiAN}$_L$ preserves the details better than other methods. Specifically, for structural images with regular buildings in Fig.~\ref{fig:Subjective comparisons:L1}, $\textit{DeFiAN}_L$ preserves the edges and repetitive structures well, and for synthesized comic scenes in Fig.~\ref{fig:Subjective comparisons:L2}, natural textures could be also preserved by using our $\textit{DeFiAN}_L$.
\begin{table*}
	\centering
    \renewcommand\arraystretch{1.15}
    \captionsetup{justification=centering}
	\caption{\textsc{\\Ablation study on DeFiAN: $\times3$ upscaling evaluation on Set5 dataset.}}
    \begin{tabular}{{c|c c c c c c c c}}
	\hline
    \hline
	{\em MSHF}&$\times$&\checkmark&$\times$&$\times$&\checkmark&\checkmark&$\times$&\checkmark\\
    {\em DiEnDec}&$\times$&$\times$&\checkmark&$\times$&\checkmark&$\times$&\checkmark&\checkmark\\
    {\em DAC}&$\times$&$\times$&$\times$&\checkmark&$\times$&\checkmark&\checkmark&\checkmark\\
    \hline
    \#Params(K)&15233.8&15291.4&15302.5&15244.1&15360.1&15301.7&15312.8&15370.4\\
    \#FLOPs(G)&290.8&291.9&292.1&290.8&293.2&291.9&292.1&293.2\\
    PSNR&33.91&34.12&34.09&33.96&34.17&34.15&34.13&{\bf34.22}\\
    \hline
    \hline
	\end{tabular}
    \vspace{-0.2cm}
	\label{tab:ablation_study}
\end{table*}

\subsubsection{Lightweight models}
Aiming at mobile applications in consumer electronics, we also propose the lightweight models with lower than 2M parameters for $\times2$, $\times3$ and $\times4$ upscaling, which are compared with several state-of-the-art lightweight SR methods, including SRCNN~\cite{DongC2016TPAMI}, FSRCNN\cite{DongC2016ECCV}, VDSR~\cite{KimJ2016CVPR_VDSR}, LapSRN~\cite{LaiW2017CVPR}, DRRN~\cite{TaiY2017CVPR}, IDN~\cite{HuiZ2018CVPR} and CARN~\cite{AhnN2018ECCV}.

As reported in TABLE~\ref{tab:full-reference_comparison_lightweight}, with comparative space and time complexities, our $\textit{DeFiAN}_S$ achieves better performances than other state-of-the-art lightweight SR methods. For example, while evaluate on Manga109, with less complexities (reduced 0.52M of parameters), our $\textit{DeFiAN}_S$ achieves 0.51dB, 0.23dB and 0.12dB higher than CARN~\cite{AhnN2018ECCV} for $\times2$, $\times3$ and $\times4$ upscaling respectively. Besides, as shown in Fig.~\ref{fig:Subjective comparisons:S1}-\ref{fig:Subjective comparisons:S2}, our $\textit{DeFiAN}_S$ preserves the details better than other state-of-the-art methods. Specifically, with fewer complexities, the existing lightweight SR models show less effectiveness on detail preservation, \eg, in Fig.~\ref{fig:Subjective comparisons:S1}, other lightweight SISR methods are failed to restore the correct mixed edges, but only our $\textit{DeFiAN}_L$ restore them well.
\begin{figure}
	\centering
	\includegraphics[width=0.95\linewidth]{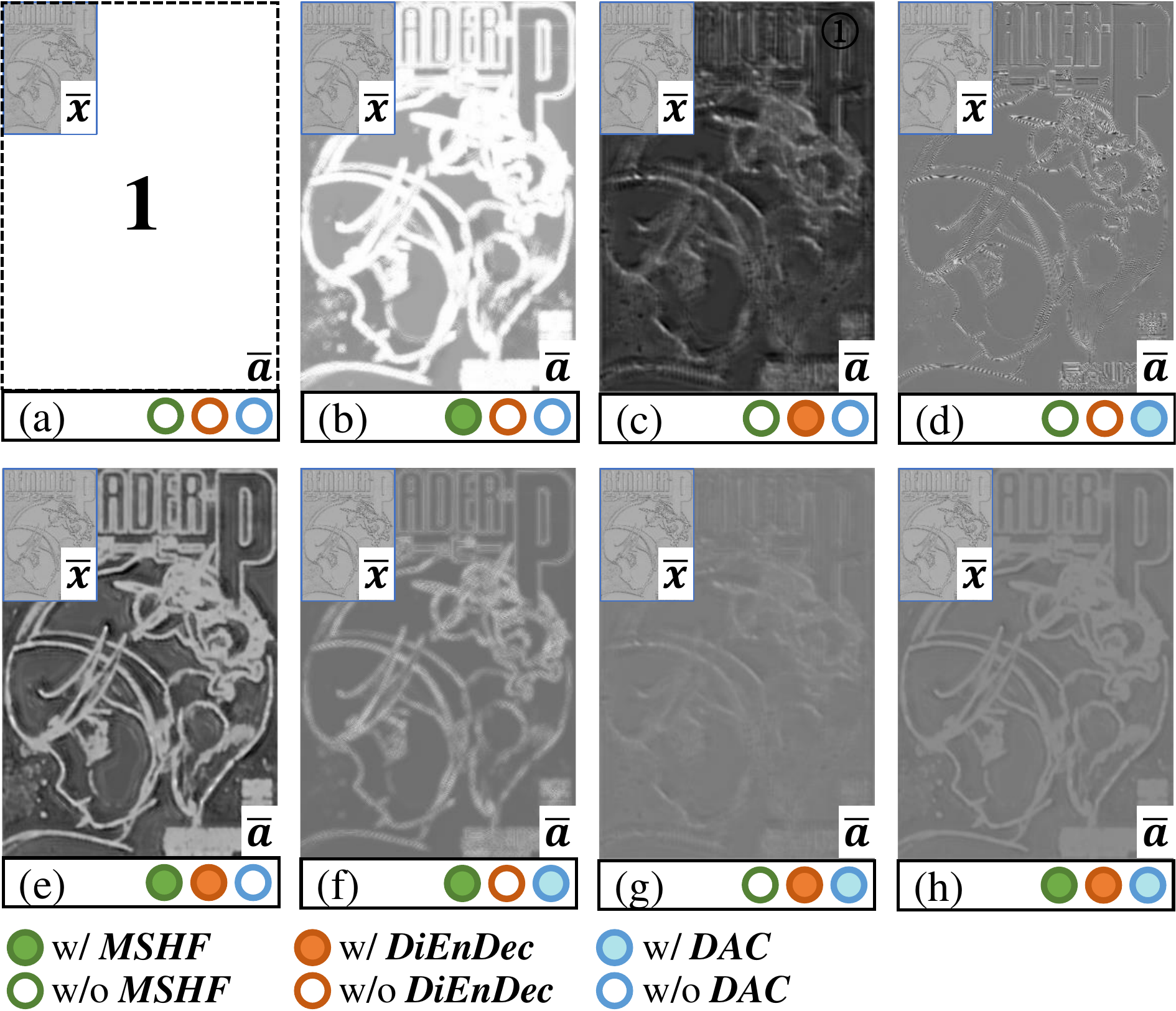}
	\caption{Visualization of ablation study on {\em DeFiAM} module.}
	\label{fig:ablation_study}
    \vspace{-0.4cm}
\end{figure}
\subsection{Model Analysis}
In this subsection, aiming at demonstrating the ability of detail-fidelity attention mechanism for interpretable detail inference in super-resolution task, the effects and contributions of each components in the proposed {\em DeFiAN} models are analyzed via experimental quantitatively verification, including the multi-scale Hessian filtering, dilated encoder-decoder, distribution alignment cell, and some implementation settings.
Particularly, the models in this subsection is set as $N=10$, $M=20$, $C=64$, and are trained with only $5\times10^4$ mini-batch updates on DIV2K datasets, unless otherwise specified.

\subsubsection{Ablation study}
Firstly, we conduct the ablation study on each component of our {\em DeFiAN} as reported in TABLE~\ref{tab:ablation_study}, including the multi-scale Hessian filtering ({\em MSHF}), dilated encoder-decoder ({\em DiEnDec}) and distribution alignment cell ({\em DAC}).
Besides, since {\em DeFiAN} is designed to focus on high-frequency details of features in network, we also visualize the intermediate features in network. As shown in Fig.~\ref{fig:ablation_study}, we visualize the averaged attention features $\boldsymbol{\bar a}$ of $\textit{DeFiAM}_{10}$ by partly removing one or more modules, including {\em MSHF}, {\em DiEnDec} and {\em DAC}:\\
$\bullet$ if remove {\em MSHF}, the attention map would suffer more undesired discrepancies in region of details, such as subfigures (e)$\rightarrow$(c) and (h)$\rightarrow$(g).\\
$\bullet$ if remove {\em DiEnDec}, the attention map would focus on the holistic map but not the details, such as subfigures (e)$\rightarrow$(b) and (h)$\rightarrow$(f).\\
$\bullet$ if remove {\em DAC}, the gaps between distributions of $\boldsymbol{x}$ and $\boldsymbol{a}$ would be enlarged which is undesired in attention operation as Eq.(\ref{eq:attention mechanism}), such as subfigures (g)$\rightarrow$(c) and (h)$\rightarrow$(e).

Note that, subfigure (a) represents $\textit{DeFiAM}$ with only {\em FEM} modules, so the averaged attention map $\boldsymbol{\bar a}$=1 and the corresponding $\textit{DeFiAN}$ is same as RCAN.
\begin{figure}
	\centering
	\includegraphics[width=1\linewidth]{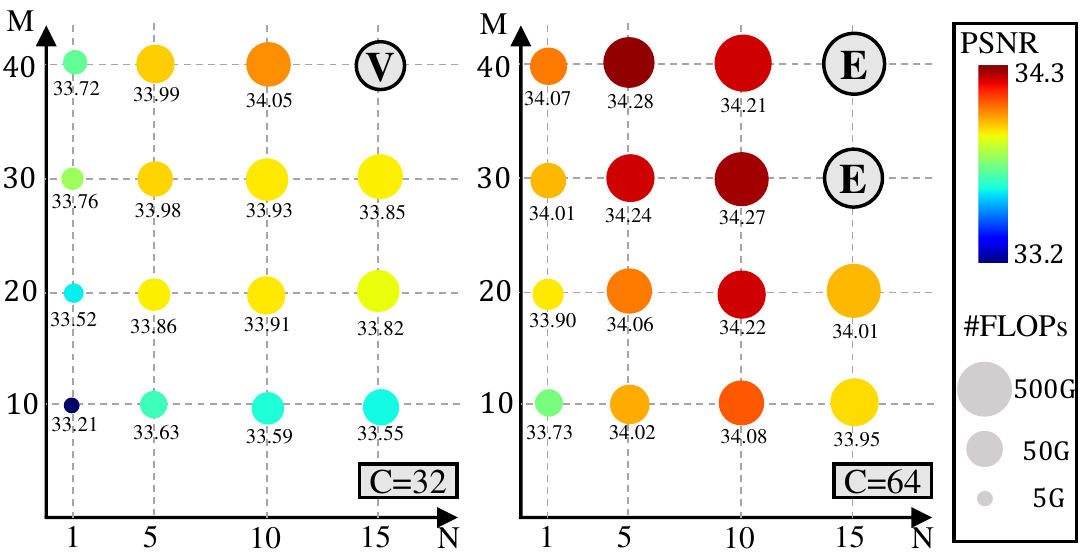}
	\caption{Comparison of {\em DeFiAN} with different model sizes: Average PSNR on Set5 dataset for $\times3$ upscaling.}
    \vspace{-0.2cm}
	\label{fig:effect_modelsize}
\end{figure}
\subsubsection{Effect of model size}
As illustrated in Section~\ref{sec:Hyperparameters}, several hyperparameters play significant roles on designing a full model of {\em DeFiAN}, including the number of {\em DeFiAM} modules $N$, the number of RCABs $M$ in each {\em DeFiAM} and the number of channels $C$ in primitive residues. We then conduct the model analysis on different settings of these hyperparameters as in Fig.~\ref{fig:effect_modelsize}. Particularly, in this subsection, all the full models consist of {\em FEM}, {\em MSHF}, {\em DiEnDec} and {\em DAC}.

In Fig.~\ref{fig:effect_modelsize}, as the model size increases (\ie, larger $N$, $M$ and $C$), the performance of {\em DeFiAN} gets a large margin of improvement but reaches the bottleneck when the model size reaches a peak level (\eg, \#FLOPs$\ge$200G) and emerges gradient exploding\textcircled{\footnotesize \bf{E}}/vanishing\textcircled{\footnotesize \bf{V}} when the depth reaches over 600 layers (the reason is discussed in Section~\ref{Sec:Discussion}).
Therefore, we choose an appropriate setting of hyperparameter as described in Section~\ref{sec:Hyperparameters} for high-fidelity and lightweight applications, which achieves excellent performance with relatively lower computational complexities.

\subsubsection{Effect of MSHF}
As aforementioned in Section~\ref{sec:Hessian filtering}, the proposed Hessian filtering is an interpretable architecture for detail representation and implemented using convolutional neural networks which is feasibly calculated in distributed GPUs. Therefore, in this subsection, we conduct some experiments to illustrate the efficiency and effect of the proposed Hessian filtering:

i) {\em Efficiency of Hessian filtering}.
In a sense, our Hessian filtering is an accelerated CNN-based algorithm for calculating the maximum eigenvalues of Hessian matrix, and faster than the pixel-wise eigenvalue solving algorithm which is formulated as
\begin{equation}
\boldsymbol{\lambda}_{klij}=max(\boldsymbol{Eig}(\boldsymbol{H}(\boldsymbol{x}_{klij})))
\label{eq:eigenvalue solver}
\end{equation}
where $\boldsymbol{H}(\boldsymbol{x}_{klij})$ represents the Hessian matrix of $klij$-th pixel using Eq.(\ref{Eq:Hessian matrix}) with 2-order gradients, $\boldsymbol{Eig}(\cdot)$ denotes the eigenvalue solver \textit{torch.eig}\footnote{\url{https://pytorch.org/docs/master/torch.html\#torch.eig}} in PyTorch.

\begin{figure}
    \centering
	\includegraphics[width=1\linewidth]{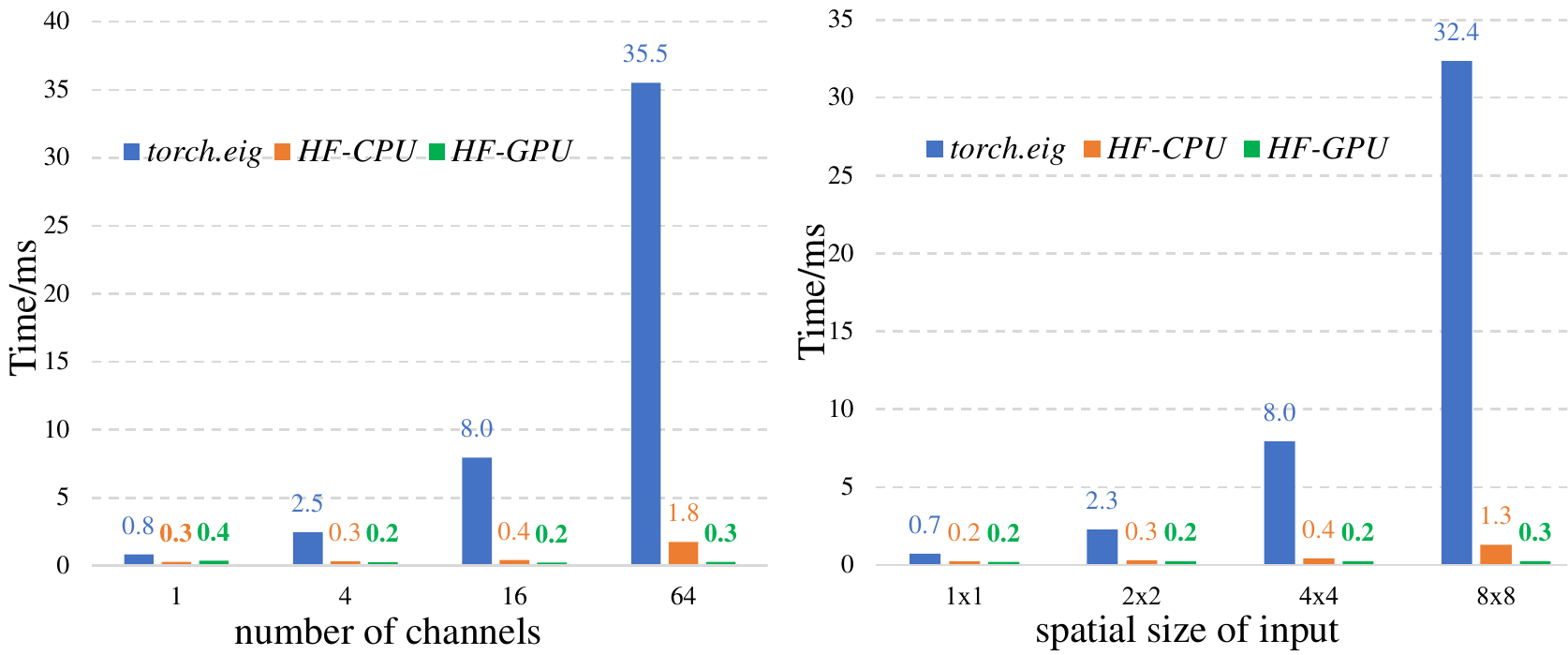}
    \caption{Comparison of Hessian filtering (CPU-based and GPU-based) and eigenvalue solver {\em torch.eig} on number of channels (spatial size is $4\times4$) and spatial size of inputs (with 16 channels).}
	\label{fig:comparison of Hessian solver}
\end{figure}

To demonstrate the superiority of our Hessian filtering, we conduct two comparative experiments to illustrate the efficiency of conventional eigenvalue solver, the proposed CPU-based and GPU-based Hessian filtering.
In particular, as Eq.(\ref{eq:eigenvalue solver}), the conventional eigenvalue solver calculates the maximum eigenvalue in pixel-wise, then for fairness, we set the input tensor of small spatial size ($1\times1$, $2\times2$, $4\times4$ and $8\times8$) and channel (1, 4, 16 and 64) to record the processing time. As shown in Fig.~\ref{fig:comparison of Hessian solver}, to process the maximum eigenvalue of a single tensor, both of our proposed CPU-based and GPU-based Hessian filterings show superiorities to the conventional solver in speed, especially with the increasing spatial sizes and channels of input, the GPU-based Hessian filtering is more robust to the size of input tensors than the others.

ii) {\em Effectiveness of Hessian filtering on multi-order gradient representation}.
Furthermore, aiming at demonstrating the effectiveness and interpretability of (multi-scale) Hessian filtering for detail representation, we conduct the visualization experiments on the inferred Hessian features and attention representations in {\em DeFiAM} modules. As Fig.~\ref{fig:visualization_hessian} shows, for the scaled Hessian filtering $\boldsymbol{\mathcal{H}}_{ker}(\cdot)$, the corresponding maximum eigenvalues of Hessian $\boldsymbol{\mathcal{\lambda}}_{ker}$ focus on detail representation with different degrees. For example on image ``{\em img086}'' in Urban100, $\boldsymbol{\mathcal{\lambda}}_{3}$ and $\boldsymbol{\mathcal{\lambda}}_{7}$ focus on the fine edges and coarse structures respectively, which  experimentally demonstrates the effectiveness of scaled Hessian filtering on representing multi-order gradients as the theoretical proof in Section~\ref{sec:Hessian filtering}.

iii) {\em Effect of multi-scale Hessian filtering}. As illustrated in Fig.~\ref{fig:MSHF} and Section~\ref{sec:MSHF}, to extract various detail-fidelity components of features, we introduce the multi-scale Hessian filtering ({\em MSHF}) to infer the multi-scale detail-fidelity attentions as Eq.(\ref{eq:MSHF_fusion}). In order to demonstrate the effect of {\em MSHF}, we compare different settings of {\em MSHF} as reported in TABLE~\ref{tab:effect_MSHF}. By applying {\em MSHF} into {\em DeFiAM} modules, the ability of network to detail preservation and representation is largely improved with 0.21dB gains on Set5 dataset. For example, {\em DeFiAM} with single scaled Hessian filtering (\eg, $ker$=$(3)$) works better than the one without any Hessian filtering, but worse than the modules with double scaled Hessian filtering (\eg, $ker$=$(3,5)$) and triple scaled Hessian filtering (\eg, $ker$=$(3,5,7)$) which performs best.
Furthermore, as in Fig.~\ref{fig:visualization_hessian}, by applying {\em MSHF} with $ker=(3,5,7)$, the detail-fidelity attention $\boldsymbol{a}$ captures both low-order and high-order details of feature and has higher ability to detail representation in field of residues.

\begin{table}
	\centering
    \captionsetup{justification=centering}
	\caption{\textsc{\\Effect of multi-scale Hessian filtering:\\Average PSNR on datasets for $\times3$ upscaling.}}
	\begin{tabular}{{p{2.4cm}<{\centering}|c|c|p{1.cm}<{\centering} p{1.cm}<{\centering}}}
	\hline
    \hline
    \multirow{2}{*}{Settings}&{\#Params}&{\#FLOPs}&\multicolumn{2}{c}{Datasets}\\
	&(K)&(G)&Set5&Manga109\\
    \hline
	w/o {\em HF}&15233.8&290.8&33.91&32.69\\
	$ker=(3)$&15253.0&291.1&34.06&32.89\\	
	$ker=(5)$&15253.0&291.1&34.05&32.85\\	
	$ker=(7)$&15253.0&291.1&33.80&32.28\\
	$ker=(3,5)$&15272.2&291.5&34.09&32.94\\	
	$ker=(3,7)$&15272.2&291.5&34.09&32.91\\
	$ker=(5,7)$&15272.2&291.5&34.04&32.91\\
	$ker=(3,5,7)$&15291.4&291.9&{\bf34.12}&{\bf33.01}\\
	\hline
    \hline
    \multicolumn{5}{l}{*$ker=(a,b,c)$ indicates the alterable kernel for MSHF.}
	\end{tabular}
	\label{tab:effect_MSHF}
\end{table}

\subsubsection{Effect of DiEnDec}
As described in Section~\ref{sec:DiEnDec}, {\em DiEnDec} is designed to meet the conditions on ``full-resolution manner'' and ``erosion \& dilation manner'' for detail-fidelity attention representation. We then conduct some experiments to demonstrate the effect of {\em DiEnDec} in these two manners:

i) {\em Full-resolution manner}. As illustrated in Fig.~\ref{fig:DiEnDec}(b), different from the existing encoder-decoder architectures (\eg, convolutional encoder-decoder ({\em ConvEnDec})~\cite{MaoX2016NIPS} which meets condition on full-resolution manner, and hourglass encoder-decoder ({\em HGEnDec})~\cite{Newell2016ECCV} which is context-aware with growing receptive fields), our {\em DiEnDec} is a full-resolution and context-aware architecture with progressively growing accumulated receptive fields. Thus, we conduct some experimental comparisons by applying different encoder-decoder (with comparative \#Params and \#FLOPs) into each {\em DeFiAM} modules as in TABLE~\ref{tab:effect_DiEnDec}.
Particularly, ``{\em w/o MSHF}" indicates the vanilla baseline model, ``{\em MSHF}" indicates the full {\em DeFiAN} model, the hyperparameter $d$ here represents the half depth of encoder-decoder (when $d$=0 and $d$=1, {\em HGEnDec}={\em ConvEnDec}={\em DiEnDec}). Compared with {\em ConvEnDec} and {\em HGEnDec}, our {\em DiEnDec} achieves better performances. Furthermore, since {\em DiEnDec} is designed for further multi-scale Hessian features fusion, the {\em DeFiAM} module with {\em MSHF} gets improvements about 0.2dB.
Besides, as the depth increases, ability of feature representation gets thrived and tends to be plateau as the balance between higher receptive field and consistency of local information.
\begin{table}
	\centering
    \renewcommand\arraystretch{1.15}
    \captionsetup{justification=centering}
	\caption{\textsc{\\Comparisons on different encoder-decoders: \\Average PSNR on Set5 dataset for $\times3$ upscaling.}} \begin{tabular}{{p{1.2cm}<{\centering}|p{.15cm}<{\centering}|p{.15cm}<{\centering}|p{0.75cm}<{\centering}|p{0.75cm}<{\centering}|p{.5cm}<{\centering}|p{.5cm}<{\centering}|p{.5cm}<{\centering}|p{.5cm}<{\centering}}}
	\hline
    \hline \multirow{3}{*}{Type}&\multirow{3}{*}{\textcircled{1}}&\multirow{3}{*}{\textcircled{2}}&\multicolumn{6}{c}{$d$ (half depth of encoder-decoder)}\\
		\cline{4-9}
		&&&\multicolumn{2}{c|}{0}&\multirow{2}{*}1&\multirow{2}{*}2&\multirow{2}{*}3&\multirow{2}{*}4\\
		\cline{4-5}
		&&&w/o {\em MSHF}&w/ {\em MSHF}&&&\\
		\hline
		{\em HGEnDec}&$\times$&\checkmark&-&-&-&34.15&34.17&34.11\\	
		{\em ConvEnDec}&$\checkmark$&$\times$&-&-&-&34.15&34.11&34.11\\	
		{\em DiEnDec}&\checkmark&\checkmark&33.96&34.15&34.13&34.19&{\bf34.22}&34.13\\	
	\hline
    \hline
		\multicolumn{9}{l}{\textcircled{1}: Characteristic of full resolution}\\
		\multicolumn{9}{l}{\textcircled{2}: Characteristic of growing receptive fields}
	\end{tabular}
	\label{tab:effect_DiEnDec}
\end{table}

\begin{figure}
	\centering
	\includegraphics[width=0.95\linewidth]{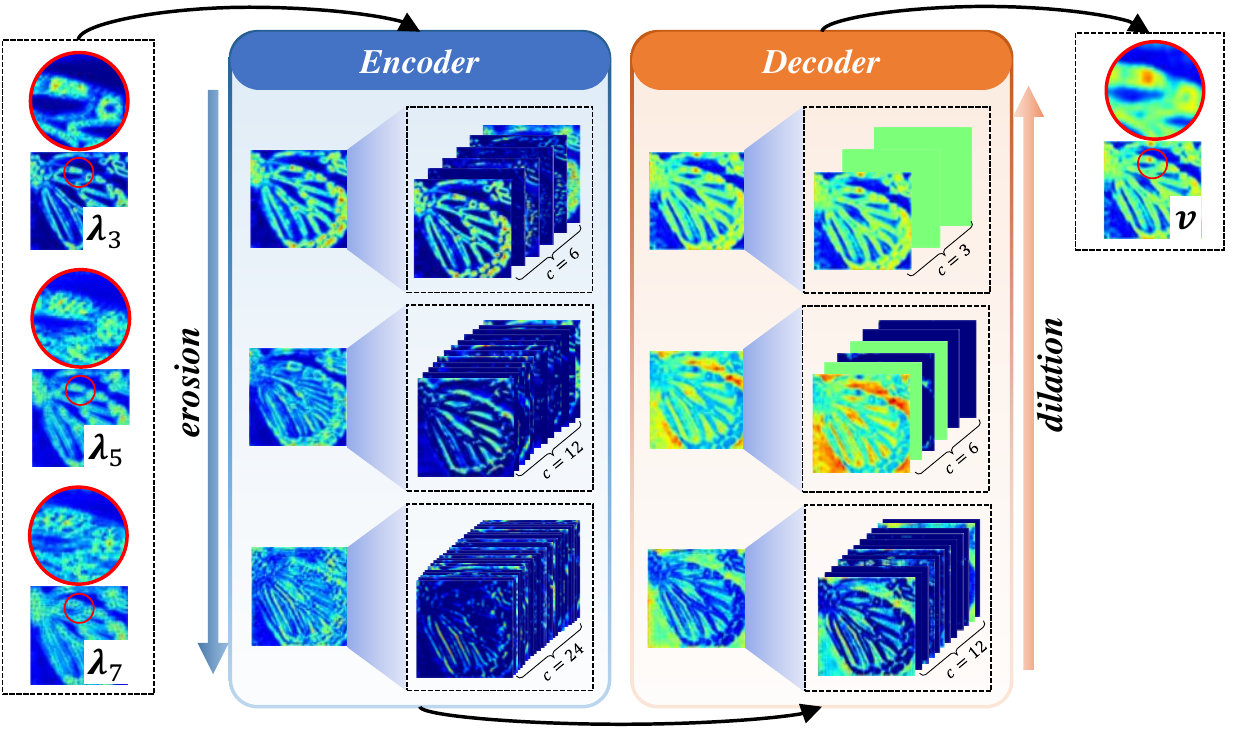}
	\caption{Visualization of {\em DiEnDec} in morphological erosion \& dilation manner. Note that, this visualization is conducted on $\textit{DeFiAM}^{(3)}$.}
	\label{fig:visualization_diendec}
\end{figure}
ii) {\em Erosion \& dilation manner}. As illustrated in Fig.~\ref{fig:DiEnDec}, for $3\times3$ kernel of dilated convolutions in {\em DiEnDec}, the accumulative receptive field for encoding and decoding could be inferred as $ARF=3\rightarrow7\rightarrow15$ and $ARF=9\rightarrow13\rightarrow15$ respectively. Moreover, morphologically, the dilated convolutions of encoder work in pixel-aggregation manner which is similar to erosion, and the dilated deconvolutions of decoder work in pixel-divergency manner which is similar to dilation. To validate the ability of {\em DiEnDec} to erosion \& dilation, we visualize the intermediate features in {\em DiEnDec} as Fig.~\ref{fig:visualization_diendec} shows. We can find that, the features are progressively fining with pixel aggregation in encoding phase and coarsening with pixel divergency in decoding phase. Namely, on the view of morphology, the encoder works for feature erosion and the decoder works for feature dilation.

\begin{figure}
	\centering
	\includegraphics[width=0.9\linewidth]{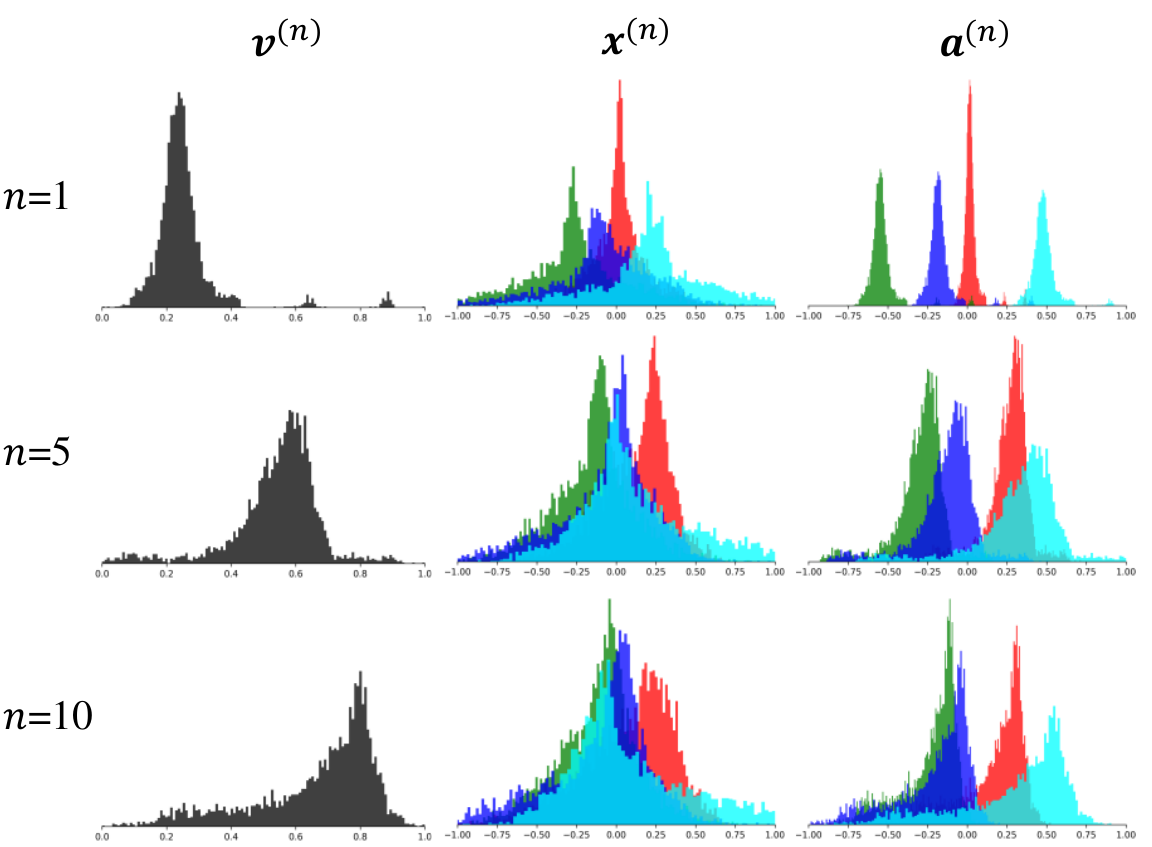}
	\caption{Visualization of {\em DAC} on distribution alignment of attention vectors. Note that, we only visualize the first 4 feature maps for presentation.}
    \vspace{-0.2cm}
	\label{fig:visualization_DAC}
\end{figure}
\subsubsection{Effect of DAC}
As aforementioned in Section~\ref{sec:DAC}, to meet the dot production and summation operations of Eq.(\ref{Eq:divide_and_conquer_eq3}), we introduce the distribution alignment cell ({\em DAC}) to align the single attention representation $\boldsymbol{v}$ on the distribution of prior features $\boldsymbol{x}$, which represent the input of {\em DeFiAM}. As shown in Fig.~\ref{fig:visualization_DAC}, the distributions of detail-fidelity attention $\boldsymbol{a}^{(n)}$ are aligned as the distributions of inputted features $\boldsymbol{x}^{(n)}$ in channel-wise and retain the spatial-wise characteristics of the inferred attention representation $\boldsymbol{v}^{(n)}$. Particularly, as in Eq.(\ref{eq:paramerization_DAC}), in order to improve the capacity of feature representation, the distribution of referenced features are transformed into $\boldsymbol{x}_r\sim \mathbb{N}(\boldsymbol{\hat \mu}_r, \boldsymbol{\hat \sigma}^2_r)$, that is why the distributions of aligned $\boldsymbol{a}^{(n)}$ are uncompletely same as $\boldsymbol{x}^{(n)}$.

\subsection{Discussion on Limitation}\label{Sec:Discussion}
In Fig.~\ref{fig:effect_modelsize}, unfortunately, as the depth of network increases over 600 layers, there exists gradient exploding/vanishing (\textcircled{\footnotesize \bf{E}}/\textcircled{\footnotesize \bf{V}}) in training phase. As illustrated in Section~\ref{sec:Hessian filtering}, for example, taking only the scales Hessian features $\boldsymbol{\lambda_3}$ into consideration, it can be inferred as an algebra combination of the second order gradient of $\boldsymbol{x}$ in Eq.(\ref{Eq: lambda_3}). Then in back propagation, the gradient of $\boldsymbol{x}$ in {\em MSHF} can be represented as
\begin{equation}
grad=\frac{\partial \boldsymbol{\mathcal{L}}}{\partial\boldsymbol{\lambda}_3}\frac{\partial \boldsymbol{\lambda}_3}{\partial\boldsymbol{x}} \propto \frac{\partial \boldsymbol{\mathcal{L}}}{\partial\boldsymbol{\lambda}_3}\frac{\partial {\nabla^2 \boldsymbol{x}}}{\partial\boldsymbol{x}} \propto \frac{\partial \boldsymbol{\mathcal{L}}}{\partial\boldsymbol{\lambda}_3}\partial^3\boldsymbol{x}
\end{equation}
where $\boldsymbol{\mathcal{L}}$ represents the loss function in Eq.(\ref{eq:objective function}). We find that the gradient of $\boldsymbol{x}$ in {\em MSHF} is proportional to the production of its 3-order partial gradients $\partial^3\boldsymbol{x}$ and $\partial \boldsymbol{\mathcal{L}}/\partial\boldsymbol{\lambda}_3$. Since $\partial \boldsymbol{\mathcal{L}}/\partial\boldsymbol{\lambda}_3$ is determined on the subsequent {\em DiEnDec}, we then set it aside. And $\partial^3\boldsymbol{x}$ is determined by only $\boldsymbol{x}$, and raises two extreme problems: (1) if $\partial^3\boldsymbol{x}\gg1$ (full of edges), the accumulated $grad$ would be exploding to $\pm\infty$; (2) if $\partial^3\boldsymbol{x}\to0$ (full of smoothes), the accumulated $grad$ would be vanishing to 0.

\section{Conclusion}

In this paper, we propose an interpretable detail-fidelity attention network for single image super-resolution, which is designed to pursue the initial intention of super-resolution for detail fidelity. Particularly, for two challenging issues on learning an adaptive operator to dividedly process low-frequency smoothes and high-frequency details, and improving the ability to detail fidelity, an interpretable multi-scale Hessian filtering is proposed, which is promising for detail fidelity representation in other computer vision applications. Besides, to improve the attention representation of Hessian features, the dilated encoder-decoder and the distribution alignment cell are proposed in morphological and statistical manners respectively. Extensive experiments demonstrate the proposed methods achieve more excellent performance than the state-of-the-art methods quantitatively and qualitatively. Moreover, we suggest that divide-and-conquer for detail fidelity is expected to be a vital issue for super-resolution in the future.


\ifCLASSOPTIONcaptionsoff
\newpage
\fi
{
	\bibliographystyle{IEEEtran}
	\bibliography{DeFiAN}

\begin{thebibliography}{10}
\providecommand{\url}[1]{#1}
\csname url@samestyle\endcsname
\providecommand{\newblock}{\relax}
\providecommand{\bibinfo}[2]{#2}
\providecommand{\BIBentrySTDinterwordspacing}{\spaceskip=0pt\relax}
\providecommand{\BIBentryALTinterwordstretchfactor}{4}
\providecommand{\BIBentryALTinterwordspacing}{\spaceskip=\fontdimen2\font plus
\BIBentryALTinterwordstretchfactor\fontdimen3\font minus
  \fontdimen4\font\relax}
\providecommand{\BIBforeignlanguage}[2]{{%
\expandafter\ifx\csname l@#1\endcsname\relax
\typeout{** WARNING: IEEEtran.bst: No hyphenation pattern has been}%
\typeout{** loaded for the language `#1'. Using the pattern for}%
\typeout{** the default language instead.}%
\else
\language=\csname l@#1\endcsname
\fi
#2}}
\providecommand{\BIBdecl}{\relax}
\BIBdecl

\bibitem{Bicubic1981TASSP}
R.~Keys, ``Cubic convolution interpolation for digital image processing,''
  \emph{IEEE Trans. Acoustics, Speech, and Signal Process.}, vol.~29, no.~6,
  pp. 1153--1160, 1981.

\bibitem{MarquinaA2008JSC}
A.~Marquina and S.~Osher, ``Image super-resolution by tv-regularization and
  bregman iteration,'' \emph{J. Sci. Comput.}, vol.~37, no.~3, pp. 367--382,
  Dec. 2008.

\bibitem{DongW2011TIP}
W.~Dong, L.~Zhang, G.~Shi, and X.~Wu, ``Image deblurring and super-resolution
  by adaptive sparse domain selection and adaptive regularization,'' \emph{IEEE
  Trans. Image Process. (TIP)}, vol.~20, no.~7, pp. 1838--1857, Jul. 2011.

\bibitem{YangJ2010TIP}
J.~Yang, J.~Wright, T.~S. Huang, and Y.~Ma, ``Image super-resolution via sparse
  representation,'' \emph{IEEE Trans. Image Process. (TIP)}, vol.~19, no.~11,
  pp. 2861--2873, 2010.

\bibitem{ZeydeR2010}
R.~Zeyde, M.~Elad, and M.~Protter, ``On single image scale-up using
  sparse-representations,'' in \emph{Curves and Surfaces}, 2010.

\bibitem{HuY2016TIP}
Y.~Hu, N.~Wang, D.~Tao, X.~Gao, and X.~Li, ``Serf: A simple, effective, robust,
  and fast image super-resolver from cascaded linear regression,'' \emph{IEEE
  Trans. Image Process. (TIP)}, vol.~25, no.~9, pp. 4091--4102, 2016.

\bibitem{HuangY2018TIP}
Y.~Huang, J.~Li, X.~Gao, L.~He, and W.~Lu, ``Single image super-resolution via
  multiple mixture prior models,'' \emph{IEEE Trans. Image Process. (TIP)},
  vol.~27, no.~12, pp. 5904--5917, 2018.

\bibitem{DongC2016TPAMI}
{C. Dong, C. C. Loy, K. He and X. Tang}, ``Image super-resolution using deep
  convolutional networks,'' \emph{IEEE Trans. Pattern Anal. Mach. Intell.
  (TPAMI)}, vol.~38, no.~2, pp. 295--307, 2016.

\bibitem{KimJ2016CVPR_VDSR}
J.~Kim, J.~K. Lee, and K.~M. Lee, ``Accurate image super-resolution using very
  deep convolutional networks,'' in \emph{IEEE Conf. Comput. Vis. Pattern
  Recognit. (CVPR)}, 2016, pp. 1646--1654.

\bibitem{HeK2016CVPR}
K.~He, X.~Zhang, S.~Ren, and J.~Sun, ``Deep residual learning for image
  recognition,'' in \emph{IEEE Conf. Comput. Vis. Pattern Recognit. (CVPR)},
  2016, pp. 1063--6919.

\bibitem{TaiY2017CVPR}
Y.~Tai, J.~Yang, and X.~Liu, ``Image super-resolution via deep recursive
  residual network,'' in \emph{IEEE Conf. Comput. Vis. Pattern Recognit.
  (CVPR)}, 2017, pp. 3147--3155.

\bibitem{LedigC2017CVPR}
C.~Ledig, L.~Theis, F.~Husz{\'{a}}r, J.~Caballero, A.~P. Aitken, A.~Tejani,
  J.~Totz, Z.~Wang, and W.~Shi, ``Photo-realistic single image super-resolution
  using a generative adversarial network,'' in \emph{IEEE Conf. Comput. Vis.
  Pattern Recognit. (CVPR)}, Jul., pp. 105--114.

\bibitem{LimB2017CVPRW}
B.~Lim, S.~Son, H.~Kim, S.~Nah, and K.~M. Lee, ``Enhanced deep residual
  networks for single image super-resolution,'' in \emph{IEEE Conf. Comput.
  Vis. Pattern Recognit. Workshops (CVPRW)}, 2017, pp. 136--144.

\bibitem{HuangG2017CVPR}
G.~Huang, Z.~Liu, L.~van~der Maaten, and K.~Q. Weinberger, ``Densely connected
  convolutional networks,'' in \emph{IEEE Conf. Comput. Vis. Pattern Recognit.
  (CVPR)}, 2017, pp. 2261--2269.

\bibitem{TongT2017ICCV}
T.~Tong, G.~Li, X.~Liu, and Q.~Gao, ``Image super-resolution using dense skip
  connections,'' in \emph{IEEE Int. Conf. Comput. Vis. (ICCV)}, 2017, pp.
  4799--4807.

\bibitem{TaiY2017ICCV}
Y.~Tai, J.~Yang, X.~Liu, and C.~Xu, ``Memnet: {A} persistent memory network for
  image restoration,'' in \emph{IEEE Int. Conf. Comput. Vis. (ICCV)}, 2017, pp.
  4549--4557.

\bibitem{ZhangY2018CVPR}
Y.~Zhang, Y.~Tian, Y.~Kong, B.~Zhong, and Y.~Fu, ``Residual dense network for
  image super-resolution,'' in \emph{IEEE Conf. Comput. Vis. Pattern Recognit.
  (CVPR)}, 2018, pp. 2472--2481.

\bibitem{HuJ2018CVPR}
J.~Hu, L.~Shen, and G.~Sun, ``Squeeze-and-excitation networks,'' in \emph{IEEE
  Conf. Comput. Vis. Pattern Recognit. (CVPR)}, 2018, pp. 7132--7141.

\bibitem{WooS2018ECCV}
S.~Woo, J.~Park, J.-Y. Lee, and I.~S. Kweon, ``Cbam: Convolutional block
  attention module,'' in \emph{Eur. Conf. Comput. Vis. (ECCV)}, 2018, pp.
  3--19.

\bibitem{WangX2018CVPR}
X.~Wang, R.~Girshick, A.~Gupta, and K.~He, ``Non-local neural networks,'' in
  \emph{IEEE Conf. Comput. Vis. Pattern Recognit. (CVPR)}, 2018, pp.
  7794--7803.

\bibitem{ZhangY2018ECCV}
Y.~Zhang, K.~Li, K.~Li, L.~Wang, B.~Zhong, and Y.~Fu, ``Image super-resolution
  using very deep residual channel attention networks,'' in \emph{Eur. Conf.
  Comput. Vis. (ECCV)}, 2018, pp. 294--310.

\bibitem{HuY2019TCSVT}
Y.~{Hu}, J.~{Li}, Y.~{Huang}, and X.~{Gao}, ``Channel-wise and spatial feature
  modulation network for single image super-resolution,'' \emph{IEEE Trans.
  Circuits Syst. Video. Technol. (TCSVT)}, pp. 1--1, 2019.

\bibitem{ZhangY2019ICLR}
\protect{Y. Zhang}, K.~Li, K.~Li, L.~Wang, B.~Zhong, and Y.~Fu, ``Residual
  non-local attention networks for image restoration,'' in \emph{Int. Conf.
  Learn. Rep. (ICLR)}, 2019.

\bibitem{KimJ2016CVPR_DRCN}
J.~Kim, J.~K. Lee, and K.~M. Lee, ``Deeply-recursive convolutional network for
  image super-resolution,'' in \emph{IEEE Conf. Comput. Vis. Pattern Recognit.
  (CVPR)}, 2016, pp. 1637--1645.

\bibitem{HuiZ2018CVPR}
Z.~Hui, X.~Wang, and X.~Gao, ``Fast and accurate single image super-resolution
  via information distillation network,'' in \emph{IEEE Conf. Comput. Vis.
  Pattern Recognit. (CVPR)}, 2018, pp. 723--731.

\bibitem{LiJ2018ECCV}
J.~Li, F.~Fang, K.~Mei, and G.~Zhang, ``Multi-scale residual network for image
  super-resolution,'' in \emph{Eur. Conf. Comput. Vis. (ECCV)}, 2018, pp.
  527--542.

\bibitem{AhnN2018ECCV}
N.~Ahn, B.~Kang, and K.-A. Sohn, ``Fast, accurate, and lightweight
  super-resolution with cascading residual network,'' in \emph{Eur. Conf.
  Comput. Vis. (ECCV)}, 2018, pp. 256--272.

\bibitem{DongC2016ECCV}
C.~Dong, C.~C. Loy, and X.~Tang, ``Accelerating the super-resolution
  convolutional neural network,'' in \emph{Eur. Conf. Comput. Vis. (ECCV)},
  2016, pp. 391--407.

\bibitem{ShiW2016CVPR}
W.~Shi, J.~Caballero, F.~Husz{\'{a}}r, J.~Totz, A.~P. Aitken, R.~Bishop,
  D.~Rueckert, and Z.~Wang, ``Real-time single image and video super-resolution
  using an efficient sub-pixel convolutional neural network,'' in \emph{IEEE
  Conf. Comput. Vis. Pattern Recognit. (CVPR)}, 2016, pp. 1874--1883.

\bibitem{LaiW2017CVPR}
W.-S. Lai, J.-B. Huang, N.~Ahuja, and M.-H. Yang, ``Deep laplacian pyramid
  networks for fast and accurate super-resolution,'' in \emph{IEEE Conf.
  Comput. Vis. Pattern Recognit. (CVPR)}, 2017, pp. 624--632.

\bibitem{HeX2019CVPR}
X.~He, Z.~Mo, P.~Wang, Y.~Liu, M.~Yang, and J.~Cheng, ``Ode-inspired network
  design for single image super-resolution,'' in \emph{IEEE Conf. Comput. Vis.
  Pattern Recognit. (CVPR)}, 2019, pp. 1732--1741.

\bibitem{Kastner2000Neuro}
S.~Kastner and L.~G. Ungerleider, ``Mechanisms of visual attention in the human
  cortex,'' \emph{Annual Review of Neuroscience}, vol.~23, pp. 315--341, Mar.
  2000.

\bibitem{Moore2017Psycho}
T.~Moore and M.~Zirnsak, ``Neural mechanisms of selective visual attention,''
  \emph{Annual Review of Psychology}, vol.~68, pp. 47--72, Jan. 2017.

\bibitem{DengH2007CVPR}
H.~Deng, W.~Zhang, E.~Mortensen, T.~Dietterich, and L.~Shapiro, ``Principal
  curvature-based region detector for object recognition,'' in \emph{IEEE Conf.
  Comput. Vis. Pattern Recognit. (CVPR)}, 2007, pp. 1--8.

\bibitem{MaoX2016NIPS}
X.-J. Mao, C.~Shen, and Y.-B. Yang, ``Image restoration using very deep
  convolutional encoder-decoder networks with symmetric skip connections,'' in
  \emph{Adv. Neural Inform. Process. Syst. (NeurIPS)}, 2016, pp. 2802--2810.

\bibitem{Newell2016ECCV}
A.~Newell, K.~Yang, and J.~Deng, ``Stacked hourglass networks for human pose
  estimation,'' in \emph{Eur. Conf. Comput. Vis. (ECCV)}, 2016, pp. 483--499.

\bibitem{HuangY2019ICME}
Y.~Huang, J.~Li, X.~Gao, W.~Lu, and Y.~Hu, ``Improving image super-resolution
  via feature re-balancing fusion,'' in \emph{IEEE Int. Conf. Multimedia and
  Expo (ICME)}, 2019, pp. 580--585.

\bibitem{Balle2017ICLR}
J.~Ball\'e, V.~Laparra, and E.~P. Simoncelli, ``End-to-end optimized image
  compression,'' in \emph{Int. Conf. Learn. Represent. (ICLR)}, 2017.

\bibitem{YuF2016ICLR}
F.~Yu and V.~Koltun, ``Multi-scale context aggregation by dilated
  convolutions,'' in \emph{Int. Conf. Learn. Represent. (ICLR)}, 2016.

\bibitem{WangP2018WACV}
P.~Wang, P.~Chen, Y.~Yuan, D.~Liu, Z.~Huang, X.~Hou, and G.~Cottrell,
  ``Understanding convolution for semantic segmentation,'' in \emph{Winter
  Conf. Appl. Comput. Vis.}, 2018.

\bibitem{BevilacC2012BMVC}
M.~Bevilacqua, A.~Roumy, C.~Guillemot, and M.-L.~A. Morel, ``Low-complexity
  single-image super-resolution based on nonnegative neighbor embedding,'' in
  \emph{British Mach. Vis. Conf. (BMVC)}, 2012.

\bibitem{ArbelaezP2011TPAMI}
P.~Arbel{\'{a}}ez, M.~Maire, C.~C. Fowlkes, and J.~Malik, ``Contour detection
  and hierarchical image segmentation,'' \emph{IEEE Trans. Pattern Anal. and
  Mach. Intell. (TPAMI)}, vol.~33, no.~5, pp. 898--916, 2011.

\bibitem{HuangJB2015CVPR}
J.-B. Huang, A.~Singh, and N.~Ahuja, ``Single image super-resolution from
  transformed self-exemplars,'' in \emph{IEEE Conf. Comput. Vis. Pattern
  Recognit. (CVPR)}, 2015, pp. 5197--5206.

\bibitem{TimofteR2017CVPRW}
R.~Timofte, E.~Agustsson, L.~V. Gool, M.-H. Yang, L.~Zhang, and et~al, ``Ntire
  2017 challenge on single image super-resolution: Methods and results,'' in
  \emph{IEEE Conf. Comput. Vis. Pattern Recognit. Workshops (CVPRW)}, 2017, pp.
  1110--1121.

\bibitem{Haris2018CVPR}
M.~Haris, G.~Shakhnarovich, and N.~Ukita, ``Deep back-projection networks for
  super-resolution,'' in \emph{IEEE Conf. Comput. Vis. Pattern Recognit.
  (CVPR)}, 2018, pp. 1664--1673.

\bibitem{KingmaD2014ICLR}
D.~P. Kingma and J.~Ba, ``Adam: {A} method for stochastic optimization,'' in
  \emph{Int. Conf. Learn. Represent. (ICLR)}, 2014.

\end{thebibliography}
}

\begin{IEEEbiography}[{\includegraphics[width=1in,height=1.25in,clip,keepaspectratio]{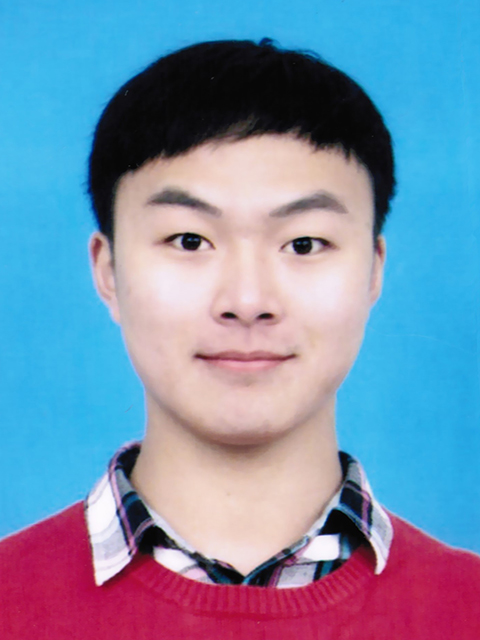}}]{Yuanfei Huang} received the B.Eng. degree in communication engineering from Hebei University of Technology, Tianjin, China, in 2016. He is currently pursuing the Ph.D. degree in circuits and systems with the Video and Image Processing System Laboratory, School of Electronic Engineering, Xidian University, Xi'an, China. His current research interests include machine learning and computer vision. In these areas, he has published technical articles in referred journals and proceedings including IEEE Transactions on Image Processing, IEEE Transactions on Circuits and Systems for Video Technology, Signal Processing and \etc.
\end{IEEEbiography}

\begin{IEEEbiography}[{\includegraphics[width=1in,height=1.25in,clip,keepaspectratio]{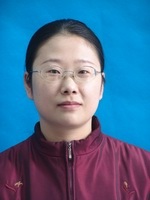}}]{Jie Li} received the B.Sc., M.Sc. and Ph.D. degrees in Circuits and Systems from Xidian University, China, in 1995, 1998 and 2005 respectively. Since 1998, she joined the School of Electronic Engineering at Xidian University. Currently, she is a Professor of Xidian University. Her research interests include computational intelligence, machine learning, and image processing. In these areas, she has published over 50 technical articles in refereed journals and proceedings including IEEE TIP, IEEE TCyb, IEEE TCSVT, Pattern Recognition etc.
\end{IEEEbiography}

\begin{IEEEbiography}[{\includegraphics[width=1in,height=1.25in,clip,keepaspectratio]{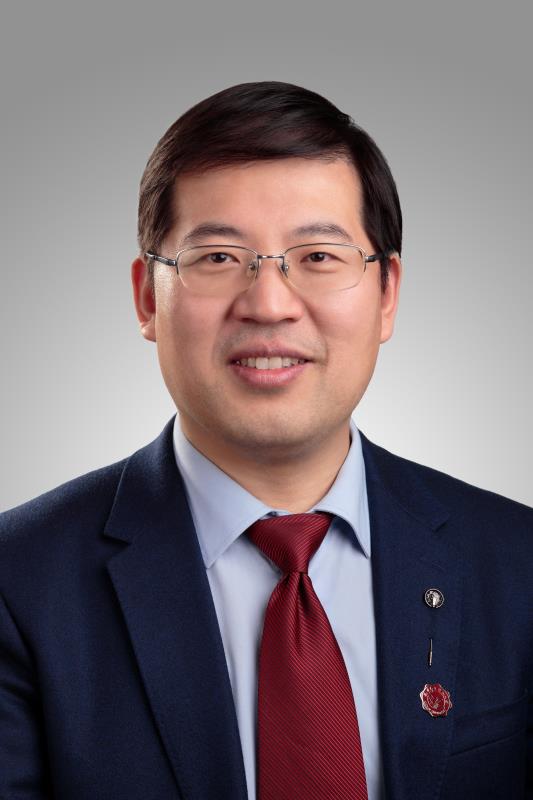}}]{Xinbo Gao}(M'02-SM'07) received the B.Eng., M.Sc., and Ph.D. degrees in signal and information processing from Xidian University, Xi'an, China, in 1994, 1997, and 1999, respectively. From 1997 to 1998, he was a Research Fellow at the Department of Computer Science, Shizuoka University, Shizuoka, Japan. From 2000 to 2001, he was a Post-doctoral Research Fellow at the Department of Information Engineering, the Chinese University of Hong Kong, Hong Kong. Since 2001, he has been at the School of Electronic Engineering, Xidian University. He is currently a Cheung Kong Professor of Ministry of Education, a Professor of Pattern Recognition and Intelligent System, and the Director of the State Key Laboratory of Integrated Services Networks, Xi'an, China. His current research interests include multimedia analysis, computer vision, pattern recognition, machine learning, and wireless communications. He has published six books and around 200 technical articles in refereed journals and proceedings. Prof. Gao is on the Editorial Boards of several journals, including Signal Processing (Elsevier) and Neurocomputing (Elsevier). He served as the General Chair/Co-Chair, Program Committee Chair/Co-Chair, or PC Member for around 30 major international conferences. He is a Fellow of the Institute of Engineering and Technology and a Fellow of the Chinese Institute of Electronics.
\end{IEEEbiography}

\begin{IEEEbiography}[{\includegraphics[width=1in,height=1.25in,clip,keepaspectratio]{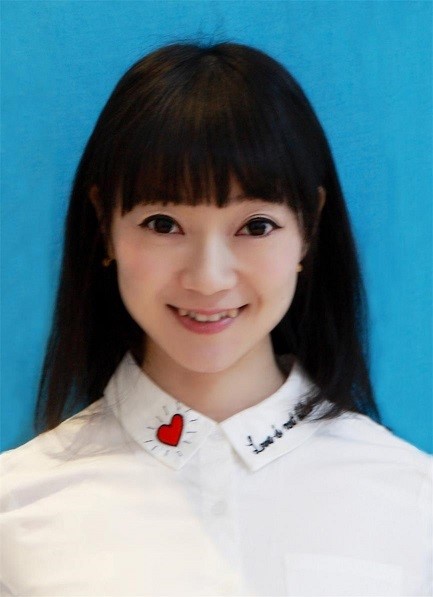}}]{Yanting Hu} received the M.Eng. degree in information and communication engineering and Ph.D. degree in pattern recognition and intelligent system from Xidian University, Xi'an, China, in 2008 and 2019, respectively. Since 2008, she has been a faculty member with the School of Medical Engineering and Technology, Xinjiang Medical University, Urumqi, China. Her current research interests include machine learning and computer vision.
\end{IEEEbiography}

\begin{IEEEbiography}[{\includegraphics[width=1in,height=1.25in,clip,keepaspectratio]{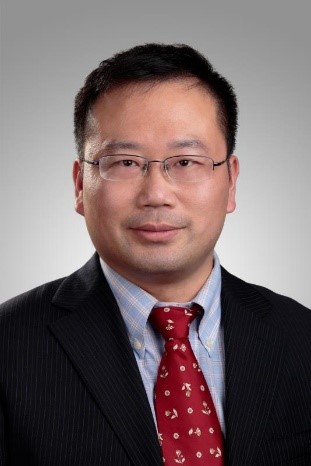}}]{Wen Lu} received the BSc, MSc and PhD degrees in signal and information processing from from Xidian University, Xi'an, China, in 2002, 2006 and 2009, respectively. Since 2009, he joined the School of School of Electronic Engineering at Xidian University. From 2010 to 2012, he was a Post-doctoral Research Fellow with the Department of Electronic Engineering, Stanford University, U.S. He is currently a professor at School of Electronic Engineering, Xidian University. His current research interests include multimedia analysis, computer vision, pattern recognition, deep learning. He has published 2 books and around 50 technical articles in refereed journals and proceedings including IEEE Transactions on Image Processing, IEEE Transactions on Cybernetics, Information Science, Neurocomputing, etc. He is also on the editorial boards and serves as reviewers for many journals, such as IEEE Transactions on Image Processing, IEEE Transactions on Multimedia, etc.
\end{IEEEbiography}

\end{document}